\documentclass[runningheads]{llncs}
\usepackage{graphicx}
\usepackage{amsmath,amssymb} 
\usepackage{color}
\usepackage{multirow}
\usepackage[table,xcdraw]{xcolor}

\begin{document}
\pagestyle{headings}
\mainmatter

\def\ACCV20SubNumber{***}  

\title{CLASS: Cross-Level Attention and Supervision for Salient Objects Detection} 
\titlerunning{CLASS: Cross-Level Attention and Supervision for Salient Objects Detection}
%

\author{Lv Tang\inst{1} \and
Bo Li\thanks{Correspondence should be addressed to Bo Li.}\inst{2}} 
\authorrunning{L.Tang and B.Li}
%
\institute{State Key Lab for Novel Software Technology, Nanjing University, Nanjing, China \and
Youtu Lab, Tencent, Shanghai, China \\
\email{luckybird1994@gmail.com}\\
\email{libraboli@tencent.com}}

\maketitle

\begin{abstract}
Salient object detection (SOD) is a fundamental computer vision task. Recently, with the revival of deep neural networks, SOD has made great progresses. However, there still exist two thorny issues that cannot be well addressed by existing methods, indistinguishable regions and complex structures. To address these two issues, in this paper we propose a novel deep network for accurate SOD, named CLASS. First, in order to leverage the different advantages of low-level and high-level features, we propose a novel non-local cross-level attention (CLA), which can capture the long-range feature dependencies to enhance the distinction of complete salient object. Second, a novel cross-level supervision (CLS) is designed to learn complementary context for complex structures through pixel-level, region-level and object-level. Then the fine structures and boundaries of salient objects can be well restored. In experiments, with the proposed CLA and CLS, our CLASS net consistently outperforms 13 state-of-the-art methods on five datasets. 
\end{abstract}

\section{Introduction}
Salient object detection (SOD) is a fundamental task in computer vision, which is derived with the goal of detecting and segmenting the most distinctive objects from visual scenes. As a preliminary step, SOD plays an essential role in various visual systems, such as object recognition \cite{DBLP:conf/cvpr/RutishauserWKP04,DBLP:journals/tcsv/RenGCT14}, semantic segmentation \cite{DBLP:journals/pami/WeiLCSCFZY17}, visual tracking \cite{DBLP:conf/icml/HongYKH15} and image-sentence matching \cite{Ji_2019_ICCV}.

\begin{figure}[t]
\centering
\includegraphics[width=0.75\columnwidth]{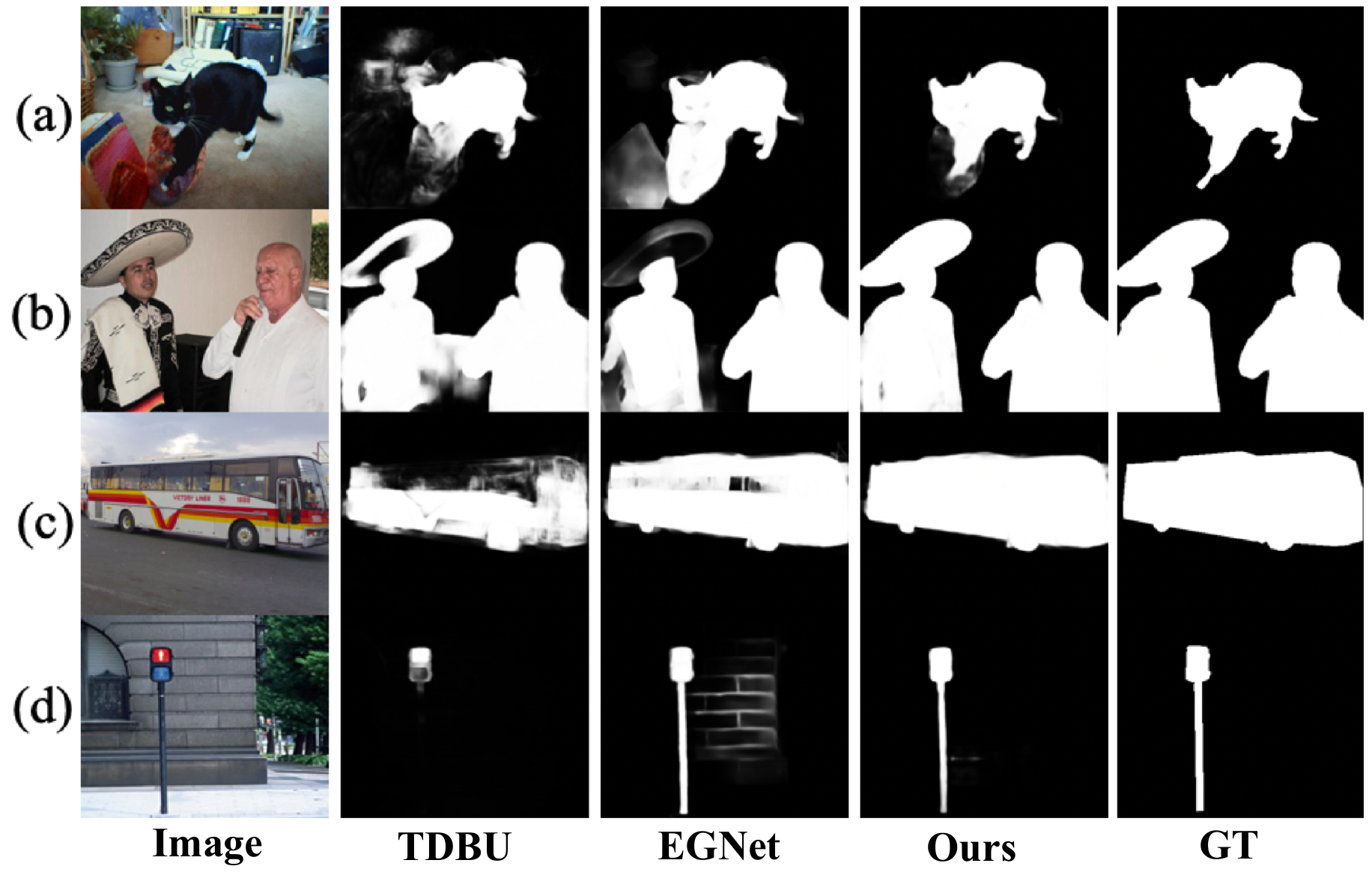}
\caption{Issues that cannot be well addressed by existing SOD methods.
In (a)(b), some ``salient-like''  regions and large appearance change between salient object parts usually confuse the models to cause wrong predictions. In (c)(d), it is hard to maintain the fine structures and boundaries of salient objects. Images
and ground-truth masks (GT) are from \cite{DBLP:conf/cvpr/LiHKRY14,DBLP:conf/cvpr/YangZLRY13}. Results are generated by
TDBU \cite{DBLP:conf/cvpr/WangSC019}, EGNet \cite{Zhao_2019_ICCV}  and our approach.}
\label{fig:example}
\end{figure}

Recently, with the application of deep convolutional neural networks (CNNs), salient object detection has achieved impressive improvements over conventional hand-crafted feature based approaches. 
Owing to their efficiency and powerful capability in visual feature representation, the CNN-based methods have pushed the performance of SOD to a new level, especially after the emergence of fully convolutional neural networks (FCNs). However, there still exist two thorny issues that cannot be well addressed by existing SOD methods. First, it is difficult to keep the uniformity and wholeness of the salient objects in some complex detecting scenes. As shown in Fig. 1(a)(b), some ``salient-like'' regions and large appearance change between salient object parts usually confuse the models to cause wrong predictions. Second, it is hard to maintain the fine structures and boundaries of salient objects(see Fig. 1(c)(d)). These two issues hinder the further development of SOD, and make it still a challenging task.  

To alleviate the first problem, some methods \cite{DBLP:journals/pami/HouCHBTT19,DBLP:conf/cvpr/WuSH19,DBLP:conf/cvpr/WangSC019,DBLP:conf/iccv/ZhangWLWR17,DBLP:conf/cvpr/LiuH018,DBLP:conf/cvpr/WangZSHB19,DBLP:conf/eccv/ChenTWH18,DBLP:conf/cvpr/FengLD19} attempt to enhance the feature by aggregating multi-level and multi-scale features  or adopting attention mechanisms to guide the models to focus on salient regions. However, these
mechanisms ignore the relationships between the object parts and the complete salient object, leading to wrong prediction in complex real-world scenarios.
For the second problem, methods \cite{DBLP:conf/eccv/LiYCLS18,Su_2019_ICCV,Zhao_2019_ICCV,DBLP:conf/cvpr/LiuHCFJ19,Wu_2019_ICCV}
try to maintain the fine structures by introducing some special boundary branch or adding extra boundary supervision. These branches can provide boundary details to restore the salient contour, but they inevitably contain some noise edges might influence the final prediction(like the bricks in Fig.1(d)). Meanwhile, these \mbox{pixel-level} boundary supervisions not only cannot capture enough context of complex structures but need extra cost to get boundary labels.

In this paper, to address the aforementioned two issues, we propose a novel convolutional neural network, named CLASS, which achieves remarkable performance in detecting accurate salient objects.
For the first issue, inspired by non-local mechanism \cite{DBLP:conf/cvpr/BuadesCM05,DBLP:conf/cvpr/0004GGH18}, we develop a novel attention module to capture the relationships between regions and the complete salient object. 
Unlike the conventional self-attention mechanism, we want to capture features dependencies through different levels, which is called cross-level attention module (CLA). On one hand, low-level features which contain the fine spatial details can guide the selection of high-level through non-local position dependencies. Thus it can assist to locate preliminary salient objects and suppress the non-salient regions. On the other hand, high-level features with rich semantic information can be used as a guidance of low-level features through channel-wise dependencies, which can keep the wholeness of salient objects with large inner appearance change. For the second issue,  in order to  restore the fine structures of salient objects, we propose a novel cross-level supervision strategy (CLS). Unlike the pixel-level boundary loss, our CLS consists of binary cross entropy loss, a novel structural similarity loss and F-measure loss, which are designed to learn complementary information from ground truth through pixel-level, region-level and object-level. These cross-level constraints can provide context of complex structures to better calibrate the saliency values.

The main contributions of this paper can be summarize as: 

(1) We propose a 
SOD network with a novel cross-level attention mechanism, which can keep the uniformity and wholeness of the detected salient objects by modeling the channel-wise and position-wise features dependencies through different levels.

(2) We introduce a novel cross-level supervision to train our network across three different levels: pixel-level, region-level and object-level. The complementarity between these losses can help
restoring the fine structures and boundaries of salient objects.

(3) We conduct comprehensive experiments on five public SOD benchmark datasets. The results demonstrate that with the above two components the proposed CLASS net consistently outperforms state-of-the-art algorithms, which proves the effectiveness and superiority of our method.

\section{Related Work}
Over the past decades, a large amount of SOD algorithms have been developed. Traditional models \cite{DBLP:journals/pami/IttiKN98,DBLP:journals/pami/ChengMHTH15,DBLP:journals/ijcv/WangJYCHZ17,DBLP:conf/eccv/WangZLSQ16,DBLP:conf/iccv/KleinF11} detect salient objects by utilizing various heuristic
saliency priors with hand-crafted features. More details about the traditional methods can be found in the survey \cite{DBLP:journals/cvm/BorjiCHJL19}. Here we mainly focus on deep learning based saliency detection models, especially the latest FCN-based methods in recent three years.

Lots of FCN-based models are devoted to exploring various feature enhancement strategies to improve the ability of localization and awareness of salient objects. Hou et al. \cite{DBLP:journals/pami/HouCHBTT19} introduced short connections to the skip-layer structures within the HED \cite{DBLP:journals/ijcv/XieT17} architecture, which provided rich multi-scale feature maps at each layer. Zhang et al. \cite{DBLP:conf/iccv/ZhangWLWR17} aggregated multi-level feature maps into multiple resolutions,  which were then fused to predict saliency maps in a recursive manner. Liu et al. \cite{DBLP:conf/cvpr/LiuH018} proposed a pixel-wise contextual attention to guide the network learning to attend global and local contexts.
Chen et al. \cite{DBLP:conf/eccv/ChenTWH18} propose a reverse
attention network, which restore the missing object parts and details by erasing the current predicted salient regions from side-output features.
 Feng et al. \cite{DBLP:conf/cvpr/FengLD19} designed the attentive feedback modules to control the message passing between encoder and decoder blocks.
 Wu et al. \cite{DBLP:conf/cvpr/WuSH19} introduced skip connection between multi-level features and a holistic attention module to refine the detection results by enlarging the coverage area of the initial saliency map.
 Wang et al. \cite{DBLP:conf/cvpr/WangSC019} proposed to integrate both top-down and bottom-up saliency inference by using multi-level features in an iterative and cooperative manner. However, these above 
mechanisms lack consideration of the relationships between the object parts and the complete salient object, leading to wrong prediction in complex \mbox{real-world} scenarios. Unlike these methods, we propose the cross-level attention module: 
the non-local position-wise and channel-wise features dependencies through different levels. The cross-level position attention can guide the network to suppress the non-salient regions, while the cross-level channel attention can
keep the wholeness of salient objects with large inner appearance change.

Recently, some methods consider leveraging boundary information to restore the fine structures of salient objects. These methods usually utilize some special boundary branch or adding extra boundary supervision to get the boundary information. 
Li et al. \cite{DBLP:conf/eccv/LiYCLS18} transferred salient knowledge from an existing contour detection model as useful priors to facilitate
feature learning in SOD. 
In \cite{Su_2019_ICCV,Zhao_2019_ICCV,Wu_2019_ICCV,DBLP:conf/cvpr/WangZSHB19}, edge features from some sophisticated edge detection branches or modules were fused with salient features as complementary information to enhance the structural details for accurate saliency detection. However, these branches inevitably contain some noise edges that might influence the final prediction(like the bricks in Fig.1(d)). Liu et al. \cite{DBLP:conf/cvpr/LiuHCFJ19} proposed to utilize extra edge supervision to jointly train an edge detection branch and a SOD branch, which can assist the deep neural network to refine the details of salient objects. 
Feng et al. \cite{DBLP:conf/cvpr/FengLD19} presented a boundary-enhanced loss as a supplement to the cross-entropy loss for learning fine boundaries.
These pixel-level boundary supervisions cannot capture enough context of complex structures and also increase labeling cost. 
Different form the above methods, our novel cross-level supervision strategy (CLS) , which consists of binary cross entropy loss, a novel structural similarity loss and F-measure loss, are designed to train our network across three different levels: pixel-level, region-level and object-level. With the learned complementary context of complex structures, it is much easier for our network to maintain the fine structures and boundaries of salient objects. For more information about the DNN-based methods, please refer to survey~\cite{DBLP:journals/corr/abs-1904-09146,DBLP:journals/spm/HanZCLX18}.

\section{Proposed Method}
In this section, we first describe the overall architecture of the proposed deep salient object detection network, and then elaborate our main contributions, which are corresponding to cross-level attention module and cross-level supervision respectively.

\begin{figure}[t]
\centering
\includegraphics[width=0.8\columnwidth]{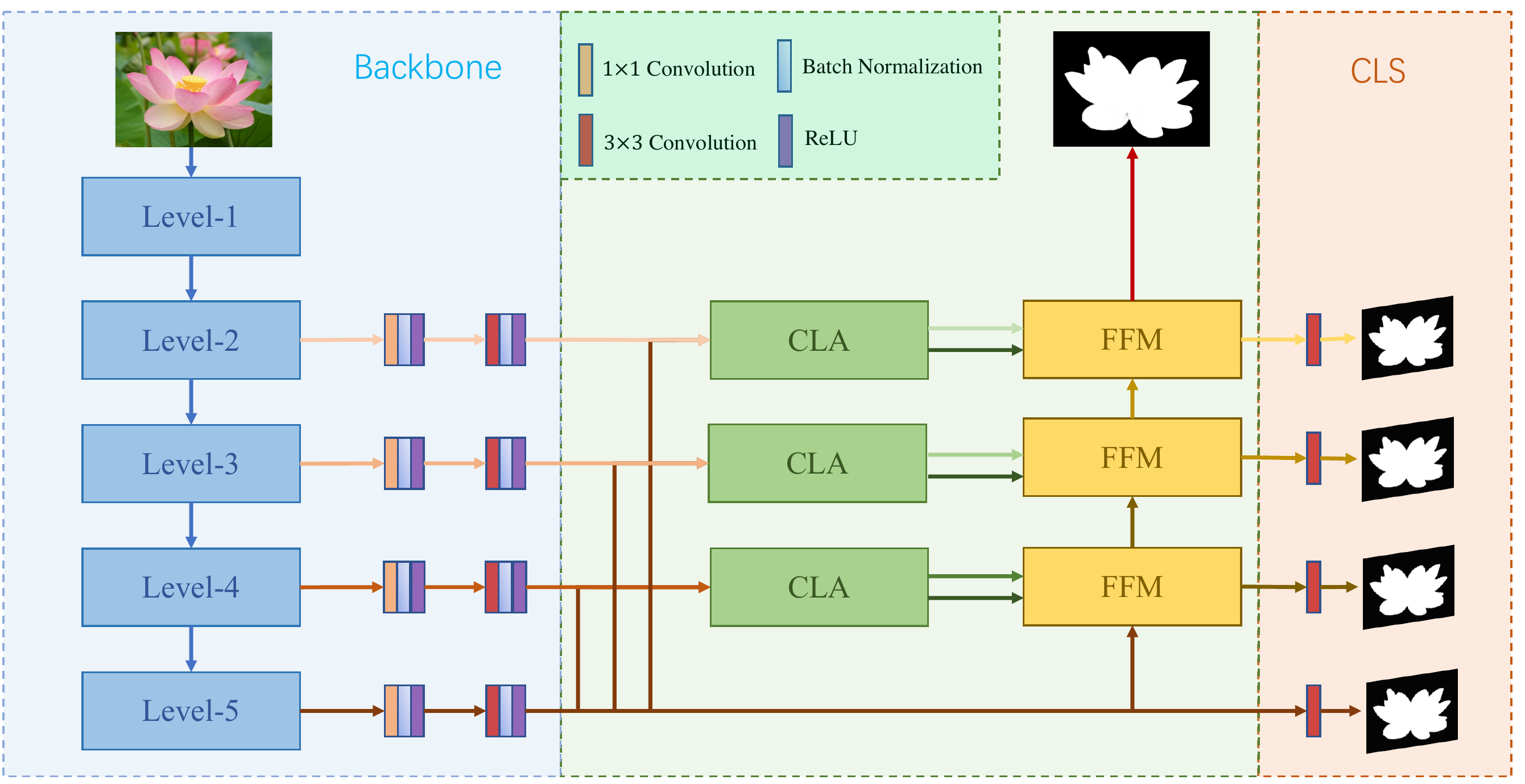}
\caption{An overview of proposed CLASS net. ResNet-50 is used as the backbone encoder. Cross-level attention module (CLA) is used to capture the long-range features dependencies between the high-level features and the low-level features.
Feature fusion module (FFM) is a basic module to fuse features for decoder. Cross-level supervision (CLS) in each stages help to ease the optimization of CLASS net.}
\label{fig:example}
\end{figure}

\subsection{Architecture}
As illustrated in Fig. 2, the proposed CLASS net has a simple U-Net-like Encoder-Decoder architecture \cite{DBLP:conf/miccai/RonnebergerFB15}.
The ResNet-50 \cite{DBLP:conf/cvpr/HeZRS16} is used as backbone feature encoder, which has five residual modules for encoding, named as level-1 to level-5 respectively. Because level-1 feature brings too much computational cost but little performance improvement, we don't use it for following process as suggested in work\cite{DBLP:conf/cvpr/WuSH19}. Between the encoder and decoder, we add two convolution blocks as the bridge. The $1\times1$ convolutional layer compresses the channels
of high-level features for subsequent processing and the $3\times3$ convolutional layer transfers features for SOD task. Each of these convolution layers is followed by a batch normalization \cite{DBLP:conf/icml/IoffeS15} and a ReLU activation \cite{DBLP:conf/nips/HahnloserS00}.  The high-level feature in level-5 is denoted as $ \{ F_{h} | h = 5 \}$, while the other three levels features are denoted as $ \{ F_{l} | l = 2,3,4 \} $. Then cross-level attention modules are used to capture the long-range features dependencies between the high-level features ($F_{h}$) and the low-level features ($F_{l}$).
For the decoder, we use a feature fusion module (FFM) to delicately aggregate the output features of CLA module in each stage and the upsampled features from the previous stage in a bottom-up manner. The output of each decoder stage is defined as $\{ D_{i} | i = 2,3,4 \}$. Cross-level supervision (CLS) is applied in each stage to train our CLASS net jointly. The output of the last stage is taken as the final saliency prediction.

\subsection{Cross-Level Attention Module}
Discriminant feature representations are essential for accurate SOD, while most existing methods cannot well keep the uniformity and wholeness of the salient objects in some complex scenes because of lacking consideration of the relationships between the indistinguishable regions and the salient object. To address this problem, inspired by non-local mechanism \cite{DBLP:conf/cvpr/BuadesCM05,DBLP:conf/cvpr/0004GGH18}, we develop a novel attention module to capture the long-range features dependencies. 
However, features in different levels usually have different recognition information. Common non-local models \cite{DBLP:conf/cvpr/0004GGH18}, which rely on a single layer feature,  
exhibit limited ability in capturing sufficient long range dependencies.
Unlike them, we want to leverage the advantages of features in different levels and propose the cross-level attention module. As illustrated in Fig. 3, we design two parts in CLA to model the channel-wise and position-wise features dependencies across the high-level feature and the low-levels features.

\begin{figure}[t]
 \centering
 \includegraphics[width=0.8\linewidth]{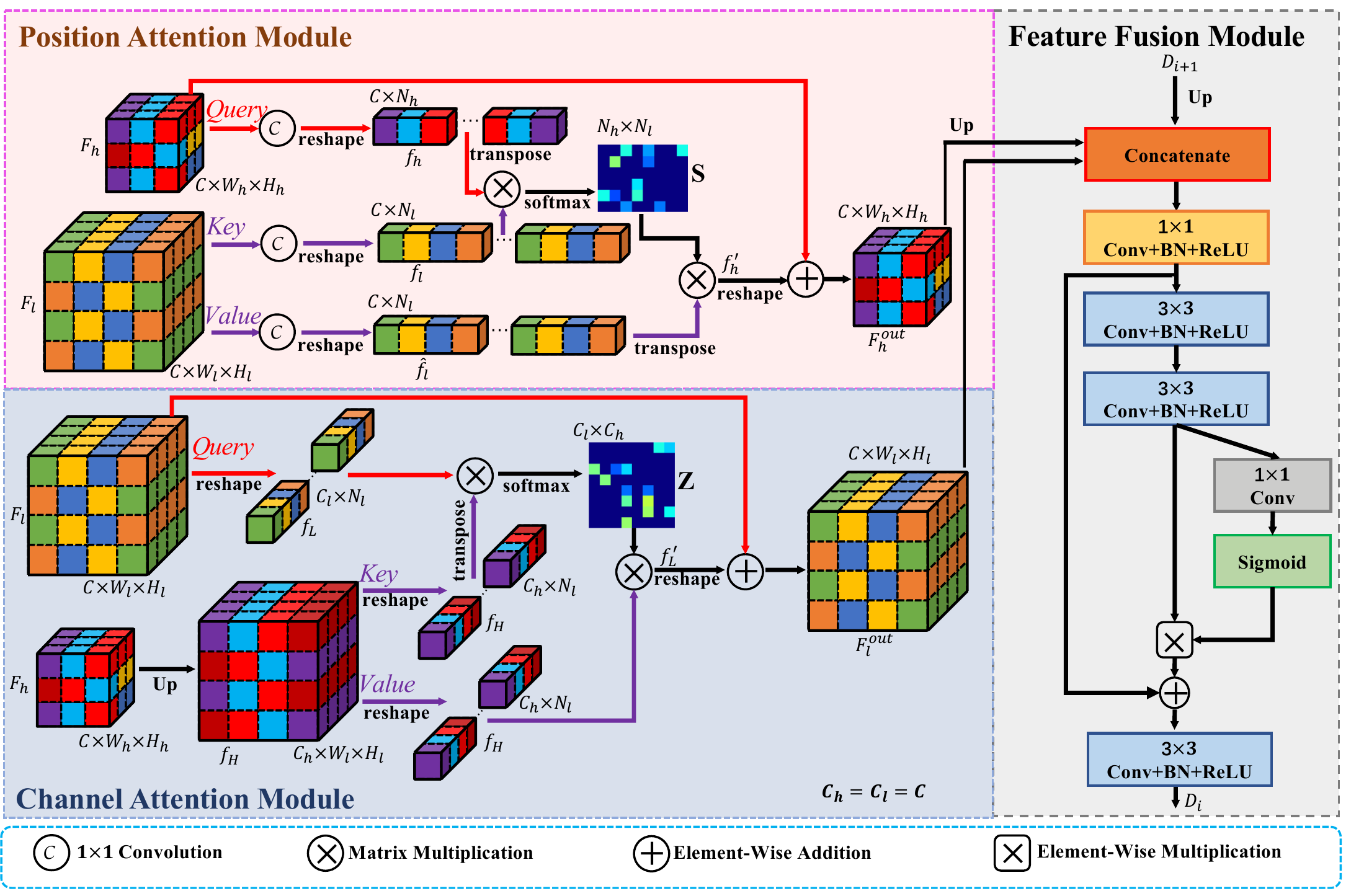}
  \caption{An overview of the proposed Cross-Level Attention Module and Feature Fusion Module. Cross-Level Attention Module contains Position Attention and Channel Attention.}
  \label{eccv_fig4}
\end{figure}

\textbf{Position Attention Module.} In some complex detecting scenes, there exist some non-salient regions which have ``salient-like''  appearance. These regions usually share some similar attributes with real salient regions like the high visual contrast. Thus, the saliency-like regions may also have high saliency semantics at the high-level layer. So the high-level feature which lacks low-level cues is difficult to distinguish saliency-like regions. We want to use the rich spatial details of low-level features as a guidance to make the high-level layer concentrate on real salient positions and then learn more discriminative features to suppress the non-salient regions. Specifically, as illustrated in Fig. 3, the input of Position Attention Module is a high-level feature map $F_h \in \mathcal{R}^{ C \times H_h \times W_h }$ and a low-level feature map $ F_l \in \mathcal{R}^{ C \times H_l \times W_l  }$. To be specific, for \textit{Query} branch, we first add a $1 \times 1$ convolution layer on $F_h$ and reshape the feature to $f_h \in \mathcal{R}^{C\times N_h}$, where $N_h = H_h \times W_h$.
Meanwhile, for \textit{Key} branch, we also use a $1 \times 1$ convolution layer on $F_l$ and reshape the feature to $f_l \in \mathcal{R}^{C\times N_l}$, where $N_l = H_l \times W_l$. After that, we perform a matrix multiplication between the transpose of $f_h$ and $f_l$, then apply a \textit{softmax} function to calculate the spatial attention map $\textbf{S} \in \mathcal{R}^{N_h \times N_l}$. Each pixel value in $\textbf{S}$ is defined as:
 \begin{equation}
S(i,j) = \frac{ exp(f_h^i \cdot f_l^j ) }{\sum_{j=1}^{N_l}exp( f_h^i \cdot f_l^j) },
\end{equation}
where $i \in [1,N_h]$, $S(i,j)$ measures the $j^{th}$ position in low-level feature impact on $i^{th}$ position in high-level feature. Meanwhile, like \textit{Key} branch, we generate feature $\hat{f}_l$ from \textit{Value} branch and perform a matrix multiplication between S and the transpose of $\hat{f}_l$ to get $f'_h \in \mathcal{R}^{C \times N_h}$, which is defined as:
 \begin{equation}
f'_h(i) = \sum_{j=1}^{N_l}S(i,j)\hat{f}_l(j),
\end{equation}
Finally, we reshape $f'_h$ to $ \mathcal{R}^{ C \times H_h \times W_h }$ and multiply it by a scale parameter $\alpha$ and perform an element-wise sum operation with $F_h$ to obtain the final output $F_h^{out} \in  \mathcal{R}^{ C \times H_h \times W_h }$. It is defined as:
 \begin{equation}
F_h^{out} = \alpha \cdot f'_h + F_h,
\end{equation}
where $\alpha$ is initialized as 0 and gradually learns to assign more weight~\cite{DBLP:conf/icml/ZhangGMO19}. 

\textbf{Channel Attention Module.} In some complicated scenarios, salient objects may have large inner appearance change. These appearance variations are mainly reflected in the difference in the channels of low-level features. Since the channel of the low-level features contains almost no semantic information but low-level visual appearance cues, it is hard to maintain the semantic consistency of the object parts. To address this issue, we want to use the rich semantics of channels in high-level features to guide the selection of low-level features, which equips our network with the power of assigning saliency label to different-looking regions to keep the wholeness of salient objects. Specifically, as illustrated in Fig. 3, for channel attention module, we first use bilinear to upsample $F_h$ to the spatial size of $F_l$, denoted as $f_H \in \mathcal{R}^{C_h \times H_l \times W_l}$, where $C_h=C$. For \textit{Query} branch, we reshape $F_l$ to $f_L \in \mathcal{R}^{C_l \times N_l} $, where $C_l=C$. For \textit{Key} branch, we reshape $f_H$ to $\mathcal{R}^{C_h \times N_l}$. Next, we perform a matrix multiplication between $f_L$ and the transpose of $f_H$ and apply a softmax function to get the channel attention map $\textbf{Z} \in \mathcal{R}^{C_l \times C_h}$. Each pixel value in $\textbf{Z}$ can be calculated as:
 \begin{equation}
Z(i,j) = \frac{ exp(f_L^i \cdot f_H^j ) }{\sum_{j=1}^{C_h}exp(f_L^i \cdot f_H^j) },
\end{equation}
where $i \in [1,C_l]$, $Z(i,j)$ measures the $j^{th}$ channel of high-level feature impact on $i^{th}$ channel of low-level feature. At the same time, for \textit{Value} branch, we reshape $f_H$ to $\mathcal{R}^{C_h \times N_l}$ and perform a matrix multiplication with $\textbf{Z}$ to get $f'_L \in \mathcal{R}^{C_l \times N_l}$, which is defined as: 
\begin{equation}
f'_L(i) = \sum_{j=1}^{C_h}Z(i,j)f_H(j),
\end{equation}
Finally, we reshape $f'_L$ to $\mathcal{R}^{C_l \times H_l \times W_l}$ and multiply it by a scale parameter $\beta$ and perform an element-wise sum operation with $F_l$ to obtain the final output $F_l^{out} \in  \mathcal{R}^{ C \times H_l \times W_l }$. It is defined as:
 \begin{equation}
F_l^{out} = \beta \cdot f'_L + F_l,
\end{equation}
where $\beta$ is initialized as 0 and gradually learns to assign more weight. 

\subsection{Feature Fusion Module}
As illustrated in Fig.3, Each decoder network stage contains feature $F_l^{out}$, $F_h^{out}$ from cross-level attention module, $D_{i+1} \in \mathcal{R}^{ C \times \frac{H_l}{2} \times \frac{W_l}{2} }$ from previous decoder network stage. As these features contain different level information, we can not simply sum up these features for decoding. Inspired by SENet~\cite{DBLP:conf/cvpr/HuSS18}, we use an attention based feature fusion module to aggregate and refine these features effectively.
Specifically, we first concatenate the three features then apply a $1 \times 1$ and two $3 \times 3$ convolutional layer with batch normalization and ReLU activation function to balance the scales of the features. Then we use a $1 \times 1$ convolutional layer and $sigmoid$ function to compute a weight map, which amounts to feature selection and combination. Finally, guided by this weight map, we can archive an effective feature representation $D_i$ for following process. Fig.3 shows the details of this design.

\subsection{Cross-Level Supervision}
Through the cross-level attention, features are enhanced for better keeping the uniformity and wholeness of the salient objects. Then we focus on  restoring the fine structures and boundaries of salient objects. Toward this end, we propose a novel cross-level supervision strategy (CLS) to learn complementary context information from ground truth through pixel-level, region-level and object-level.

Let $\mathcal{I}=\{I_{n}\}^N_{n=1}$ and their groundtruth  $\mathcal{G} = \{G_{n}\}^N_{n=1}$ denote a collection of training samples where $N$ is the number of training images. After saliency detection, saliency maps are  $\mathcal{S} = \{S_{n}\}^N_{n=1}$. In SOD, binary cross entropy (BCE) is the most widely used loss function, and it is a pixel-wise loss which is defined as:
 \begin{equation}
L_{Pixel} = -\big(\ G_nlog(S_n) + (1-G_n)log(1-S_n) \ \big).
\end{equation}
From the formula of BCE loss, we find that it only considers the independent relationship between each pixel, which cannot capture  enough  context  of  complex structures, leading to blurry boundaries.

To address this problem, we propose to model region-level similarity  as a supplement to the pixel-level constraint. Following the setting of ~\cite{DBLP:journals/tip/WangBSS04,DBLP:conf/iccv/FanCLLB17}, we use the sliding window fashion to generate two corresponding regions from saliency map $S_n$ and groundtruth $G_n$, denoted as $S_n^{region} = \{S_n^i:i=1,...M\}$ and $G_n^{region} = \{G_n^i:i=1,...M\}$, where $M$ is the total number of region. Then, we adopt the simplified 2-Wasserstein distance\cite{DBLP:journals/corr/BerthelotSM17,DBLP:journals/pami/HeWST19} to evaluate the distributional similarity between $S_n^i$ and $G_n^i$. Thus the proposed network can be trained by minimizing the similarity distance $SSD_i$ between the corresponding regions, which is defined as:
\begin{equation}
SSD_i = || \mu_{S_n^i} - \mu_{G_n^i} ||_2^2 + || \sigma_{S_n^i} - \sigma_{G_n^i} ||_2^2,
\end{equation}
where local statistics $\mu_{S_n^i}$, $\sigma_{S_n^i}$ is mean and std vector of $S_n^i$, $\mu_{G_n^i}$, $\sigma_{G_n^i}$ is mean and std vector of $G_n^i$. Finally, the overall loss function is defined as:
\begin{equation}
L_{Region} = \frac{1}{M}\sum_{i=1}^{M}SSD_i, 
\end{equation}

Pixel-level and region-level constraints can only capture local context for 
salient objects, a global constraint is still needed for accurate SOD.
F-measure is often used to measure the overall similarity between the  saliency map of the detected object and its groundtruth~\cite{DBLP:conf/cvpr/MargolinZT14,DBLP:journals/tip/BorjiCJL15,DBLP:conf/cvpr/WangSDB18}. Hence we want to directly optimize the F-measure to learn the global information, called object-level supervision. For easy remembering, we denote F-measure as $F_\beta$ in the following. 
The predicted saliency map $S_n$ is a non-binary map, so we calculate $F_\beta$ value via two steps. First, multiple thresholds are applied to the predicted saliency map to obtain multiple binary maps. Then, these binary maps are compared to the groundtruth. 
Hence, the whole process of calculating $F_\beta$ is nondifferentiable. However, we can modify it to be differentiable. 
Considering pixel value $G_{n}(x,y)$ and $S_{n}(x,y)$, if $G_{n}(x,y) = 1$ and $S_{n}(x,y) = p$, it means this pixel has $p$ probability to be true positive and $(1-p)$ probability to be false negative; if $G_{n}(x,y) = 0$ and $S_{n}(x,y) = p$, it means this pixel has $p$ probability to be true negative and $1-p$ to be false positive. So, we can calulate precision and recall by following Formulation:
\begin{equation}
precision = \frac{TP}{TP+FP} = \frac{S_n \cdot G_n}{S_n \cdot G_n + S_n \cdot (1-G_n) } = \frac{S_n \cdot G_n}{S_n+\epsilon},
\end{equation}
\begin{equation}
recall = \frac{TP}{TP+FN} = \frac{S_n \cdot G_n}{S_n \cdot G_n + (1-S_n) \cdot G_n } = \frac{S_n \cdot G_n}{G_n+\epsilon},
\end{equation}
\begin{equation}
F_\beta = \frac{ (1+\beta^2) \cdot precision \cdot recall }{ \beta^2 \cdot precision + recall },
\end{equation}
where $\cdot$ means pixel-wise multiplication, $\epsilon=1e^{-7}$ is a regularization constant to avoid division of zeros. $L_{Object}$ loss function is defined as:
\begin{equation}
L_{Object} = 1 -  F_\beta.
\end{equation}
Note that all parts of our network are trained jointly, and the over all loss function is given as:
\begin{equation}
L = L_{Object} + L_{Region} + L_{Pixel} .
\end{equation}

In addition, as show in Fig. 2, we use multi-level supervision as an as an auxiliary loss to facilitate sufficient training. The network has $K$ levels and the whole loss is defined as:
\begin{equation}
L_{Final} = \sum_{i=1}^{K=4}\frac{1}{2^{i-1}}L_{i} .
\end{equation}
In this loss function, high level loss has smaller weight because of its larger error. Finally, these cross-level constraints can provide context of complex structures to better calibrate the saliency values.

\section{Experiments}
\subsection{Implementation Details}
Following the works\cite{Su_2019_ICCV,Wu_2019_ICCV,Zhao_2019_ICCV,DBLP:conf/cvpr/WuSH19}, we train our proposed network on DUTS-TR.  ResNet-50~\cite{DBLP:conf/cvpr/HeZRS16} is used as the backbone network. For a more comprehensive demonstration, we also trained our network with VGG-16~\cite{DBLP:journals/corr/SimonyanZ14a} backbone.
The whole network is trained end-to-end by stochastic gradient descent(SGD). Maximum learning rate is set to 0.005 for ResNet-50 or VGG-16 backbone and 0.05 for other parts. Warm-up and linear decay strategies are used to adjust the learning rate. Momentum and weight decay are set to 0.9 and 0.0005. Batchsize is set to 32 and maximum epoch is set to 100. We use Pytorch\footnote{https://pytorch.org/} to implement our model. Only horizontal flip and multi-scale input images are utilized for data augmentation as done in \cite{Su_2019_ICCV,Wu_2019_ICCV,DBLP:conf/eccv/ChenTWH18,DBLP:conf/cvpr/FengLD19}.
A RTX 2080Ti GPU is used for acceleration. During testing, 
the proposed method runs at about 40 fps
with about $352 \times 352$ resolution without any post-processing.
Our code has been released
\footnote{https://github.com/luckybird1994/classnet}.

We comprehensively evaluated our method on five representative datasets,
including HKU-IS \cite{DBLP:conf/cvpr/LiY15}, ECSSD \cite{DBLP:journals/pami/ShiYXJ16}, PASCAL-S \cite{DBLP:conf/cvpr/LiHKRY14}, DUT-OMRON \cite{DBLP:conf/cvpr/YangZLRY13} and DUTS \cite{DBLP:conf/cvpr/WangLWF0YR17}, which contain 4447, 1000, 850, 5168 and
5019 images respectively. All datasets are human-labeled with pixel-wise ground-truth. Among them, more recent datasets PASCAL-S and DUT-TE are more challenging with salient objects that have large appearance change and complex background.

\subsection{Evaluation Metrics} 
To evaluate the performance of the proposed method, four widely-used metrics are adopted: (1) Precision-Recall (PR) curve, which shows the tradeoff between precision and recall for different threshold (ranging from 0 to 255). (2) F-measure, ($F_{\beta}$), a weighted mean of average precision and average recall, calculated by
$F_{\beta}=\frac{(1+\beta^{2}) \times Precision \times Recall}{\beta^{2} \times Precision + Recall}$. We set $\beta^{2}$ to be 0.3 as suggested in \cite{DBLP:journals/tip/BorjiCJL15}. (3) Mean Absolute Error (MAE), which characterize the average $1$-norm distance between ground truth maps and predictions. (4) Structure Measure ($S_m$), a metric to evaluate the spatial structure similarities of saliency maps based on both region-aware structural similarity $S_{r}$ and object-aware structural similarity $S_{o}$, defined as $S_{\alpha} = \alpha \ast S_{r} + (1- \alpha) \ast S_{o}$, where $\alpha = 0.5$~\cite{DBLP:conf/iccv/FanCLLB17}.

\begin{table}[htbp]
\caption{Performance of 13 state-of-the-arts and the proposed method on five benchmark datasets. Smaller MAE, larger $F_{\beta}$ and $S_m$
correspond to better performance. The best results of different backbones are in \textcolor{blue}{blue} and \textcolor{red}{red} fonts. "$\dagger$" means the results are post-processed by dense conditional random field(CRF)~\cite{DBLP:conf/nips/KrahenbuhlK11}. MK: MSRA10K~\cite{DBLP:journals/pami/ChengMHTH15}, DUTS: DUTS-TR~\cite{DBLP:conf/cvpr/WangLWF0YR17}, MB: MSRA-B~\cite{DBLP:journals/pami/LiuYSWZTS11}. }
\centering
\scalebox{0.6}{
\begin{tabular}{ccccccccccccccccc}
\hline
\multicolumn{1}{c|}{}                         & \multicolumn{1}{c|}{}                                                                             & \multicolumn{3}{c|}{ECSSD}                                                                                          & \multicolumn{3}{c|}{DUTS-TE}                                                                                        & \multicolumn{3}{c|}{DUT-OMRON}                                                                                      & \multicolumn{3}{c|}{PASCAL-S}                                                                                       & \multicolumn{3}{c}{HKU-IS}                                                                    \\ \cline{3-17}
\multicolumn{1}{c|}{\multirow{-2}{*}{Models}} & \multicolumn{1}{c|}{\multirow{-2}{*}{\begin{tabular}[c]{@{}c@{}}Training\\ dataset\end{tabular}}} & \multicolumn{1}{c|}{$F_{\beta}$} & \multicolumn{1}{c|}{$S_m$}   & \multicolumn{1}{c|}{MAE}                          & \multicolumn{1}{c|}{$F_{\beta}$} & \multicolumn{1}{c|}{$S_m$}   & \multicolumn{1}{c|}{MAE}                          & \multicolumn{1}{c|}{$F_{\beta}$} & \multicolumn{1}{c|}{$S_m$}   & \multicolumn{1}{c|}{MAE}                          & \multicolumn{1}{c|}{$F_{\beta}$} & \multicolumn{1}{c|}{$S_m$}   & \multicolumn{1}{c|}{MAE}                          & \multicolumn{1}{c|}{$F_{\beta}$} & \multicolumn{1}{c|}{$S_m$}   & MAE                          \\ \hline
\multicolumn{17}{c}{VGG-16 backbone}                                                                                                                                                                                                                                                                                                                                                                                                                                                                                                                                                                                                                                                                                                       \\ \hline
\multicolumn{1}{c|}{Amulet(ICCV2017) \cite{DBLP:conf/iccv/ZhangWLWR17}}         & \multicolumn{1}{c|}{MK}                                                                           & 0.868                            & 0.894                        & \multicolumn{1}{c|}{0.059}                        & 0.678                            & 0.804                        & \multicolumn{1}{c|}{0.085}                        & 0.647                            & 0.781                        & \multicolumn{1}{c|}{0.098}                        & 0.757                            & 0.814                        & \multicolumn{1}{c|}{0.097}                        & 0.841                            & 0.886                        & 0.051                        \\
\multicolumn{1}{c|}{C2SNet(ECCV2018) \cite{DBLP:conf/eccv/LiYCLS18}}         & \multicolumn{1}{c|}{MK}                                                                           & 0.853                            & 0.882                        & \multicolumn{1}{c|}{0.059}                        & 0.710                            & 0.817                        & \multicolumn{1}{c|}{0.066}                        & 0.664                            & 0.780                        & \multicolumn{1}{c|}{0.079}                        & 0.754                            & 0.821                        & \multicolumn{1}{c|}{0.085}                        & 0.839                            & 0.873                        & 0.051                        \\
\multicolumn{1}{c|}{RAS(ECCV2018) \cite{DBLP:conf/eccv/ChenTWH18}}            & \multicolumn{1}{c|}{MB}                                                                           & 0.889                            & 0.893                        & \multicolumn{1}{c|}{0.056}                        & 0.751                            & 0.839                        & \multicolumn{1}{c|}{0.059}                        & 0.713                            & 0.814                        & \multicolumn{1}{c|}{0.062}                        & 0.777                            & 0.792                        & \multicolumn{1}{c|}{0.101}                        & 0.871                            & 0.887                        & 0.045                        \\
\multicolumn{1}{c|}{PiCA-V(CVPR2018) \cite{DBLP:conf/cvpr/LiuH018}}         & \multicolumn{1}{c|}{DUTS}                                                                         & 0.885                            & 0.914                        & \multicolumn{1}{c|}{0.046}                        & 0.749                            & 0.861                        & \multicolumn{1}{c|}{0.054}                        & 0.710                            & 0.826                        & \multicolumn{1}{c|}{0.068}                        & 0.789                            & 0.842                        & \multicolumn{1}{c|}{0.077}                        & 0.870                            & 0.906                        & 0.042                        \\
\multicolumn{1}{c|}{DSS$\dagger$(TPAMI2019)  \cite{DBLP:journals/pami/HouCHBTT19}}           & \multicolumn{1}{c|}{MB}                                                                           & 0.904                            & 0.882                        & \multicolumn{1}{c|}{0.052}                        & 0.808                            & 0.820                        & \multicolumn{1}{c|}{0.057}                        & 0.740                            & 0.790                        & \multicolumn{1}{c|}{0.063}                        & 0.801                            & 0.792                        & \multicolumn{1}{c|}{0.093}                        & 0.902                            & 0.878                        & 0.040                        \\
\multicolumn{1}{c|}{PAGE(CVPR2019) \cite{DBLP:conf/cvpr/WangZSHB19}}           & \multicolumn{1}{c|}{MK}                                                                           & 0.906                            & 0.912                        & \multicolumn{1}{c|}{0.042}                        & 0.777                            & 0.854                        & \multicolumn{1}{c|}{0.052}                        & 0.736                            & 0.824                        & \multicolumn{1}{c|}{0.062}                        & 0.806                            & 0.835                        & \multicolumn{1}{c|}{0.075}                        & 0.882                            & 0.903                        & 0.037                        \\
\multicolumn{1}{c|}{AFNet(CVPR2019) \cite{DBLP:conf/cvpr/FengLD19}}          & \multicolumn{1}{c|}{DUTS}                                                                         & 0.908                            & 0.913                        & \multicolumn{1}{c|}{0.042}                        & 0.792                            & 0.867                        & \multicolumn{1}{c|}{0.046}                        & 0.738                            & 0.826                        & \multicolumn{1}{c|}{{\color[HTML]{3531FF} 0.057}} & 0.820                            & 0.848                        & \multicolumn{1}{c|}{0.070}                        & 0.888                            & 0.905                        & 0.036                        \\
\multicolumn{1}{c|}{CPD-V(CVPR2019) \cite{DBLP:conf/cvpr/WuSH19}}          & \multicolumn{1}{c|}{DUTS}                                                                         & 0.915                            & 0.910                        & \multicolumn{1}{c|}{0.040}                        & 0.813                            & 0.867                        & \multicolumn{1}{c|}{0.043}                        & 0.745                            & 0.818                        & \multicolumn{1}{c|}{{\color[HTML]{3531FF} 0.057}} & 0.820                            & 0.838                        & \multicolumn{1}{c|}{0.072}                        & 0.896                            & 0.904                        & 0.033                        \\
\multicolumn{1}{c|}{TSPOA(ICCV2019) \cite{Liu_2019_ICCV}}       & \multicolumn{1}{c|}{DUTS}                                                                         & 0.900                            & 0.907                        & \multicolumn{1}{c|}{0.046}                        & 0.776                            & 0.860                        & \multicolumn{1}{c|}{0.049}                        & 0.716                            & 0.818                        & \multicolumn{1}{c|}{0.061}                        & 0.803                            & 0.836                        & \multicolumn{1}{c|}{0.076}                        & 0.882                            & 0.902                        & 0.038                        \\
\multicolumn{1}{c|}{BANet-V(ICCV2019)  \cite{Su_2019_ICCV}}        & \multicolumn{1}{c|}{DUTS}                                                                         & 0.910                            & 0.913                        & \multicolumn{1}{c|}{0.041}                        & 0.789                            & 0.861                        & \multicolumn{1}{c|}{0.046}                        & 0.731                            & 0.819                        & \multicolumn{1}{c|}{0.061}                        & 0.812                            & 0.834                        & \multicolumn{1}{c|}{0.078}                        & 0.887                            & 0.902                        & 0.037                        \\
\multicolumn{1}{c|}{EGNet-V(ICCV2019) \cite{Zhao_2019_ICCV}}        & \multicolumn{1}{c|}{DUTS}                                                                         & 0.913                            & 0.913                        & \multicolumn{1}{c|}{0.041}                        & 0.800                            & 0.878                        & \multicolumn{1}{c|}{0.044}                        & 0.744                            & 0.813                        & \multicolumn{1}{c|}{{\color[HTML]{3531FF} 0.057}} & 0.809                            & 0.837                        & \multicolumn{1}{c|}{0.076}                        & 0.893                            & 0.910                        & 0.035                        \\
\multicolumn{1}{c|}{Ours}                     & \multicolumn{1}{c|}{DUTS}                                                                         & {\color[HTML]{3531FF} 0.917}     & {\color[HTML]{3531FF} 0.915} & \multicolumn{1}{c|}{{\color[HTML]{3531FF} 0.038}} & {\color[HTML]{3531FF} 0.833}     & {\color[HTML]{3531FF} 0.880} & \multicolumn{1}{c|}{{\color[HTML]{3531FF} 0.039}} & {\color[HTML]{3531FF} 0.749}     & {\color[HTML]{3531FF} 0.820} & \multicolumn{1}{c|}{{\color[HTML]{3531FF} 0.057}} & {\color[HTML]{3531FF} 0.838}     & {\color[HTML]{3531FF} 0.853} & \multicolumn{1}{c|}{{\color[HTML]{3531FF} 0.062}} & {\color[HTML]{3531FF} 0.909}     & {\color[HTML]{3531FF} 0.915} & {\color[HTML]{3531FF} 0.031} \\ \hline
\multicolumn{17}{c}{ResNet50 backbone}                                                                                                                                                                                                                                                                                                                                                                                                                                                                                                                                                                                                                                                                                                     \\ \hline
\multicolumn{1}{c|}{PiCA-R(CVPR2018) \cite{DBLP:conf/cvpr/LiuH018}}         & \multicolumn{1}{c|}{DUTS}                                                                         & 0.886                            & 0.917                        & \multicolumn{1}{c|}{0.046}                        & 0.759                            & 0.869                        & \multicolumn{1}{c|}{0.051}                        & 0.717                            & 0.832                        & \multicolumn{1}{c|}{0.065}                        & 0.792                            & 0.848                        & \multicolumn{1}{c|}{0.074}                        & 0.870                            & 0.904                        & 0.043                        \\
\multicolumn{1}{c|}{TDBU(CVPR2019) \cite{DBLP:conf/cvpr/WangSC019}}           & \multicolumn{1}{c|}{MK}                                                                           & 0.880                            & 0.918                        & \multicolumn{1}{c|}{0.041}                        & 0.767                            & 0.865                        & \multicolumn{1}{c|}{0.048}                        & 0.739                            & 0.837                        & \multicolumn{1}{c|}{0.061}                        & 0.775                            & 0.844                        & \multicolumn{1}{c|}{0.070}                        & 0.878                            & 0.907                        & 0.038                        \\
\multicolumn{1}{c|}{CPD-R(CVPR2019) \cite{DBLP:conf/cvpr/WuSH19}}          & \multicolumn{1}{c|}{DUTS}                                                                         & 0.917                            & 0.918                        & \multicolumn{1}{c|}{0.037}                        & 0.805                            & 0.869                        & \multicolumn{1}{c|}{0.043}                        & 0.747                            & 0.825                        & \multicolumn{1}{c|}{0.056}                        & 0.820                            & 0.842                        & \multicolumn{1}{c|}{0.070}                        & 0.891                            & 0.905                        & 0.034                        \\
\multicolumn{1}{c|}{SCRN(ICCV2019) \cite{Wu_2019_ICCV}}           & \multicolumn{1}{c|}{DUTS}                                                                         & 0.918                            & 0.927                        & \multicolumn{1}{c|}{0.037}                        & 0.808                            & 0.885                        & \multicolumn{1}{c|}{0.040}                        & 0.746                            & 0.837                        & \multicolumn{1}{c|}{0.056}                        & 0.827                            & 0.848                        & \multicolumn{1}{c|}{0.062}                        & 0.896                            & 0.916                        & 0.034                        \\
\multicolumn{1}{c|}{BANet(ICCV2019) \cite{Su_2019_ICCV}}          & \multicolumn{1}{c|}{DUTS}                                                                         & 0.923                            & 0.924                        & \multicolumn{1}{c|}{0.035}                        & 0.815                            & 0.879                        & \multicolumn{1}{c|}{0.040}                        & 0.746                            & 0.832                        & \multicolumn{1}{c|}{0.059}                        & 0.823                            & 0.845                        & \multicolumn{1}{c|}{0.069}                        & 0.900                            & 0.913                        & 0.032                        \\
\multicolumn{1}{c|}{EGNet(ICCV2019) \cite{Zhao_2019_ICCV}}          & \multicolumn{1}{c|}{DUTS}                                                                         & 0.920                            & 0.925                        & \multicolumn{1}{c|}{0.037}                        & 0.815                            & 0.887                        & \multicolumn{1}{c|}{0.039}                        & 0.755                            & 0.837                        & \multicolumn{1}{c|}{0.053}                        & 0.817                            & 0.846                        & \multicolumn{1}{c|}{0.073}                        & 0.901                            & 0.918                        & 0.031                        \\
\multicolumn{1}{c|}{Ours}                     & \multicolumn{1}{c|}{DUTS}                                                                         & {\color[HTML]{FE0000} 0.933}     & {\color[HTML]{FE0000} 0.928} & \multicolumn{1}{c|}{{\color[HTML]{FE0000} 0.033}} & {\color[HTML]{FE0000} 0.856}     & {\color[HTML]{FE0000} 0.894} & \multicolumn{1}{c|}{{\color[HTML]{FE0000} 0.034}} & {\color[HTML]{FE0000} 0.774}     & {\color[HTML]{FE0000} 0.838} & \multicolumn{1}{c|}{{\color[HTML]{FE0000} 0.052}} & {\color[HTML]{FE0000} 0.849}     & {\color[HTML]{FE0000} 0.863} & \multicolumn{1}{c|}{{\color[HTML]{FE0000} 0.059}} & {\color[HTML]{FE0000} 0.921}     & {\color[HTML]{FE0000} 0.923} & {\color[HTML]{FE0000} 0.028} \\ \hline
\end{tabular}}
\end{table}

\begin{figure}[htp]
\centering
\includegraphics[width=0.75\columnwidth]{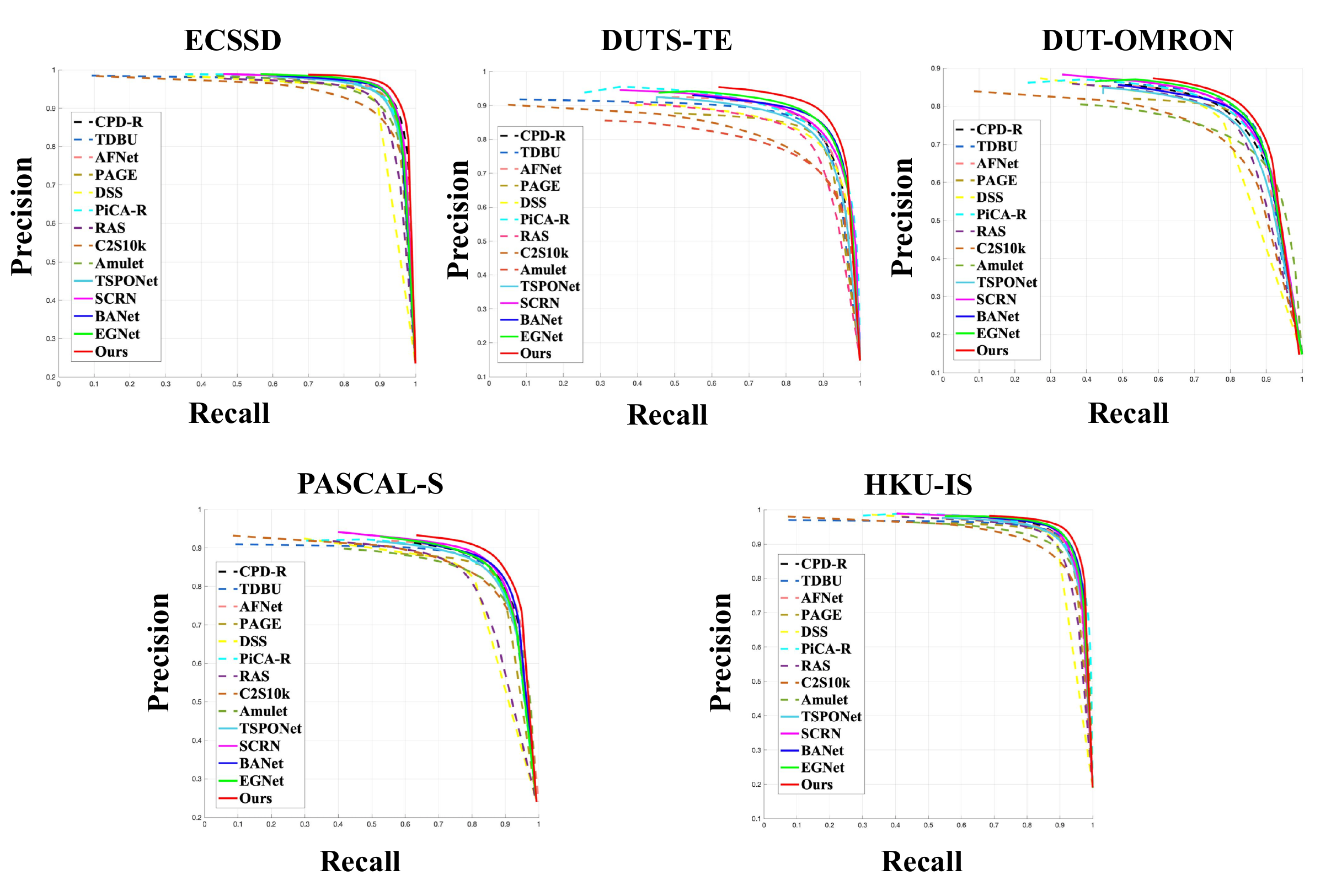}
\caption{Comparison of the PR curves across five benchmark datasets.}
\label{fig:example}
\end{figure}

\subsection{Comparisons with the State-of-the-Arts}
We compare our approach CLASS net with 13
state-of-the-art methods, including Amulet \cite{DBLP:conf/iccv/ZhangWLWR17}, C2SNet \cite{DBLP:conf/eccv/LiYCLS18}, RAS \cite{DBLP:conf/eccv/ChenTWH18}, PiCAnet \cite{DBLP:conf/cvpr/LiuH018}, DSS \cite{DBLP:journals/pami/HouCHBTT19},  PAGE \cite{DBLP:conf/cvpr/WangZSHB19}, AFNet \cite{DBLP:conf/cvpr/FengLD19}, CPD \cite{DBLP:conf/cvpr/WuSH19}, TSPOANet \cite{Liu_2019_ICCV}, TDBU \cite{DBLP:conf/cvpr/WangSC019}, SCRN \cite{Wu_2019_ICCV}, BANet \cite{Su_2019_ICCV} and EGNet \cite{Zhao_2019_ICCV}. For fair comparison, we obtain the saliency maps of these methods from authors or the deployment codes provided by authors.

\textbf{Quantitative Evaluation.} The proposed approach is
compared with 13 state-of-the-art SOD methods on five datasets, and the results are reported in Table 1 and Fig. 4.
From Table 1, we can see that our method consistently outperforms other methods across all the five benchmark datasets. It is noteworthy that our method improves the F-measure and S-measure achieved by the best-performing existing algorithms by a large margin on two challenging datasets PASCAL-S ($F_{\beta}$: 0.849 against 0.827, $S_{m}$: 0.863 against 0.848) and DUTS-TE ($F_{\beta}$: 0.856 against 0.815, $S_{m}$: 0.894 against 0.887). As for MAE, our method obviously exceed other state-of-the-art algorithms on all five datasets. When using VGG-16 as backbone, our method still consistently outperfrom other methods,  which verifies that our proposed CLA and CLS can achieve great performance with different backbone. For overall comparisons, PR curves of different methods are displayed in Fig. 4. One can observe that our approach noticeably higher than all the other methods. These observations present
the efficiency and robustness of our CLASS net across various challenging datasets, which indicates that the perspective of CLA for the problem of SOD is useful.

\textbf{Qualitative Evaluation.} To exhibit the superiority of the proposed approach, Fig. 5 show representative examples of saliency maps generated by our approach and other state-of-the-art algorithms. 
As can be seen, the proposed method can keep the uniformity and wholeness of the salient objects meanwhile maintain the fine structures and boundaries in various challenging scenes. From the column of 1 and 2 in Fig. 5, we can observe that with the influence of ``salient-like''  regions (mountain and water reflection), existing methods usually give wrong predictions. While, in our method, by the guidance of position-wise cross-level attention, the salient objects are accurately located and the non-salient regions are well suppressed. Example in third column with large inner appearance change can cause incomplete detection problem in existing methods. With the help of channel-wise cross-level attention, our method can better keep the wholeness of the salient object.
Moreover, for the case of multiple and small objects in the of 4 to 6, our method can detect all the salient objects with the relationship information captured by cross-level attention, whereas the other methods mostly miss objects or introduce some background noise.  From the column of 7 and 8, we can find that most existing methods cannot maintain the fine structures and boundaries of objects in the case of low contrast between salient object and background as well as the complicated scene. 
Note that some methods with special edge branches (EGNet, BANet and SCRN) can keep some structural details of example in column 8 and 9. However, These branches inevitably contain some noise edges can introduce background noise in the final prediction. 
It can be clearly observed that our method achieves impressive performance in all these cases, which indicates the effectiveness of cross-level supervision in maintaining the fine structures and boundaries of salient objects.

\begin{figure}[hbt!]
\centering
\includegraphics[width=0.72\columnwidth]{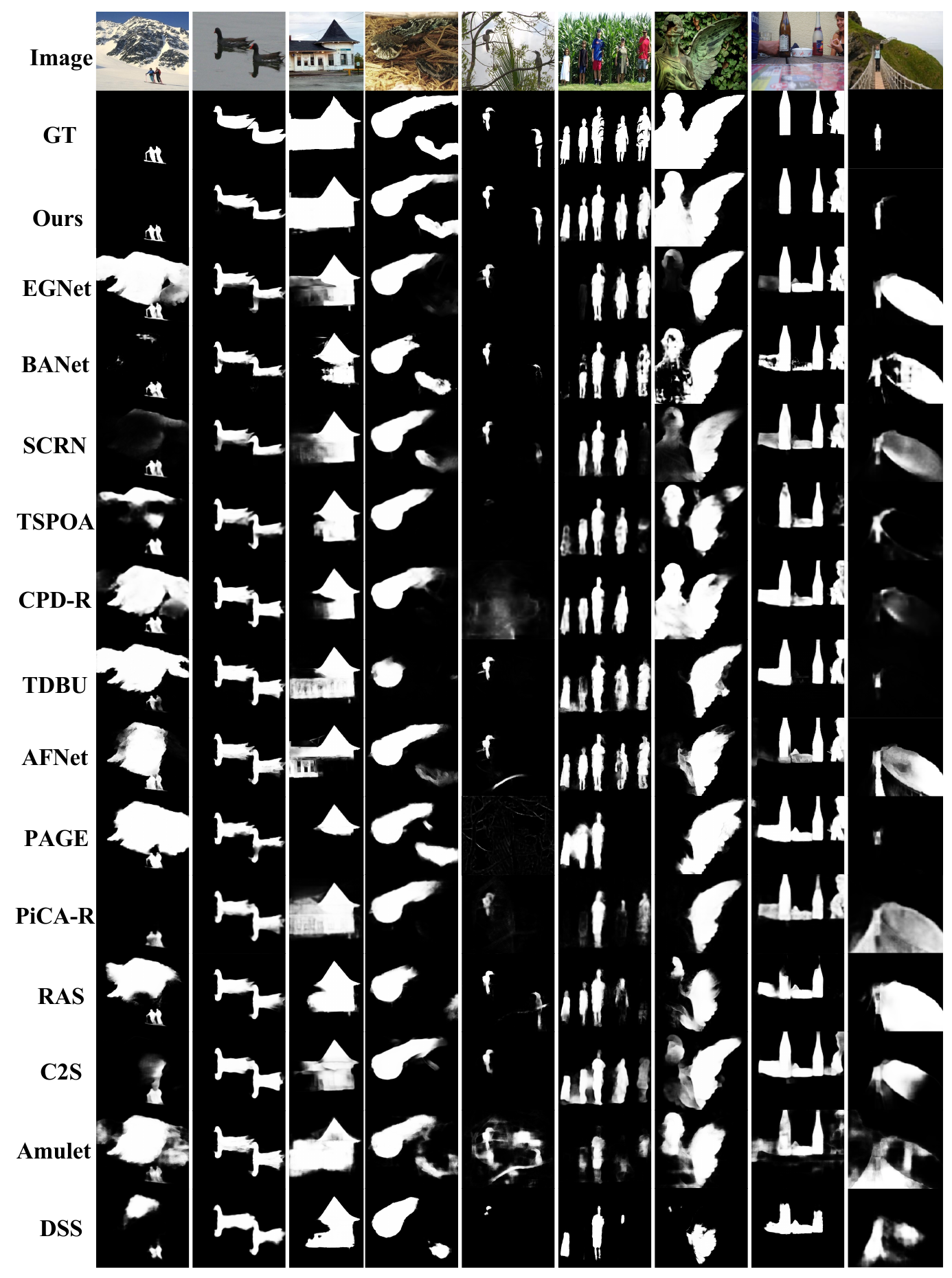}
\caption{Qualitative comparisons of the state-of-the-art algorithms and our approach.}
\end{figure}

\begin{table}[htbp]
\centering
\caption{Ablation study on different settings of supervision and architecture.}
\scalebox{0.6}{
\begin{tabular}{c|c|ccccccc|ccc|ccc|ccc}
\hline
\multirow{2}{*}{Ablation}     & \multicolumn{8}{c|}{Configurations}                                                                                                                                                                                   & \multicolumn{3}{c|}{ECSSD}  & \multicolumn{3}{c|}{PASCAL-S} & \multicolumn{3}{c}{DUT-TE} \\ \cline{2-18} 
                              &     & \begin{tabular}[c]{@{}c@{}}Pixel\\ Level\end{tabular} & \begin{tabular}[c]{@{}c@{}}Region\\ Level\end{tabular} & \begin{tabular}[c]{@{}c@{}}Object\\ Level\end{tabular} & MS      & CLA-C   & CLA-P   & FFM     & $F_{\beta}$ & $S_m$ & MAE   & $F_{\beta}$  & $S_m$  & MAE   & $F_{\beta}$ & $S_m$ & MAE   \\ \hline
\multirow{4}{*}{Loss}         & 1.  & $\surd$                                               &                                                        &                                                        &         &         &         &         & 0.900       & 0.910 & 0.043 & 0.808        & 0.838  & 0.072 & 0.782       & 0.873 & 0.044 \\
                              & 2.  & $\surd$                                               & $\surd$                                                &                                                        &         &         &         &         & 0.915       & 0.918 & 0.040 & 0.825        & 0.846  & 0.068 & 0.820       & 0.882 & 0.039 \\
                              & 3.  & $\surd$                                               & $\surd$                                                & $\surd$                                                &         &         &         &         & 0.918       & 0.919 & 0.039 & 0.832        & 0.850  & 0.067 & 0.837       & 0.885 & 0.038 \\
                              & 4.  & $\surd$                                               & $\surd$                                                & $\surd$                                                & $\surd$ &         &         &         & 0.920       & 0.920 & 0.038 & 0.838        & 0.853  & 0.064 & 0.841       & 0.887 & 0.037 \\ \hline
\multirow{6}{*}{Architecture} & 5.  & $\surd$                                               & $\surd$                                                & $\surd$                                                & $\surd$ &         &         & $\surd$ & 0.923       & 0.922 & 0.037 & 0.842        & 0.855  & 0.062 & 0.845       & 0.888 & 0.036 \\
                              & 6.  & $\surd$                                               & $\surd$                                                & $\surd$                                                & $\surd$ & $\surd$ & $\surd$ &         & 0.930       & 0.926 & 0.034 & 0.847        & 0.860  & 0.060 & 0.852       & 0.892 & 0.035 \\
                              & 7.  & $\surd$                                               & $\surd$                                                & $\surd$                                                & $\surd$ & $\surd$ &         & $\surd$ & 0.927       & 0.924 & 0.036 & 0.845        & 0.859  & 0.062 & 0.850       & 0.889 & 0.036 \\
                              & 8.  & $\surd$                                               & $\surd$                                                & $\surd$                                                & $\surd$ &         & $\surd$ & $\surd$ & 0.926       & 0.925 & 0.036 & 0.844        & 0.858  & 0.061 & 0.851       & 0.891 & 0.035 \\
                              & 9.  & $\surd$                                               &                                                        &                                                        & $\surd$ & $\surd$ & $\surd$ & $\surd$ & 0.919       & 0.925 & 0.037 & 0.830        & 0.861  & 0.063 & 0.813       & 0.891 & 0.039 \\
                              & 10. & $\surd$                                               & $\surd$                                                & $\surd$                                                & $\surd$ & $\surd$ & $\surd$ & $\surd$ & 0.933       & 0.928 & 0.033 & 0.849        & 0.863  & 0.059 & 0.856       & 0.894 & 0.034 \\ \hline
                              
\end{tabular}}
\end{table}

\subsection{Ablation Studies}
To validate the effectiveness of the proposed components of our method, we conduct a series of experiments on three datasets with different settings.

\textbf{Supervision ablation.} To investigate the effectiveness of our
proposed cross-level supervision, we conduct a set of experiments over different losses based on a baseline U-Net architecture. 
As listed in Table 2, we can observe a remarkable and consistent improvement brought by different level supervisions. Compared with only using pixel-level supervision, adding region-level structural similarity supervision can significantly improve the performance on all three metrics, especially the S-measure, which shows its ability of maintaining fine structures and boundaries of salient objects. Object-level supervision further improve the performance on F-measure. When these supervision are combined and applied at each stage (MS), we can get the best SOD results. In addition, by comparing setting No.9 and No.10, we can find that CLS is still useful even when the results is advanced.

\textbf{Architecture ablation.} To prove the effectiveness of our CLASS net, we report the quantitative comparison results of our model with different architectures. 
As shown in Table 2, Comparing No.5 and No.4, only using FFM can slightly improve the performance. Comparing No.6 and No.4, More significant improvements can be observed when we add channel-wise cross-level attention(CLA-C) and position-wise cross-level attention(CLA-P). Comparing No.7 with No.5, or No.8 with No.5, independently using CLA-C or CLA-P can also improve the performance. Finally, a best performance can be achieved through the combination of the CLA and FFM compared with baseline architecture(No.4), which verifies the compatibility of the two attentions and effectiveness of the features fusion module. For more comprehensive analyses of our proposed method, please refer to the supplementary materials.

\section{Conclusions}
In this paper, we revisit the two thorny issues that hinder
the development of salient object detection.  The issues 
consist of indistinguishable regions and complex structures.
To address these two issues, in this paper we propose a novel deep network for accurate SOD, named CLASS. For the first issue, we propose a novel non-local cross-level attention (CLA), which can leverage the advantages of features in different levels to capture the long-range feature dependencies. With the guidance of the relationships between low-level and high-level features, our model can better keep the uniformity and wholeness of the salient objects in some complex scenes. For the second issue, a novel cross-level supervision (CLS) is designed to learn complementary context for complex structures through pixel-level, region-level and object-level. Then the fine structures and boundaries of salient objects can be well restored. Extensive experiments on five benchmark datasets have validated the effectiveness of the proposed approach.

\clearpage

\title{Supplementary Materials for CLASS: Cross-Level Attention and Supervision for Salient Objects Detection}
\titlerunning{CLASS: Cross-Level Attention and Supervision for Salient Objects Detection}
%

\author{Lv Tang\inst{1} \and
Bo Li\thanks{Correspondence should be addressed to Bo Li.}\inst{2}} 
\authorrunning{L.Tang and B.Li}
%
\institute{State Key Lab for Novel Software Technology, Nanjing University, Nanjing, China \and
Youtu Lab, Tencent, Shanghai, China \\
\email{luckybird1994@gmail.com}\\
\email{libraboli@tencent.com}}
\maketitle

\section{Introduction}
This supplemental material contains three parts:

\begin{itemize}
\item Section 2 gives more quantitative and qualitative experimental results to compare our CLASS net with the state-of-the-art methods.

\item Section 3 gives an investigation of failure cases. 

\item Section 4 provides more comprehensive analyses of the proposed cross-level attention and cross-level supervision to further demonstrate the novelty of our method.
\end{itemize}
We hope this supplemental material can help you get a better understanding of our work.

\section{More Quantitative and Qualitative Comparison}
Due to the limitation of the paper length, we provide more quantitative and qualitative experimental results in this section.
\subsection{Qualitative Comparison}
As shown in Fig.1, we provide a comprehensive qualitative
comparison of our method with other 13 methods on challenging cases.
These visual examples can further demonstrate that our method is able
to handle various challenging cases and produce accurate
salient objects with high quality structure details.

\begin{figure}[ht!]
\centering
\includegraphics[width=1\columnwidth]{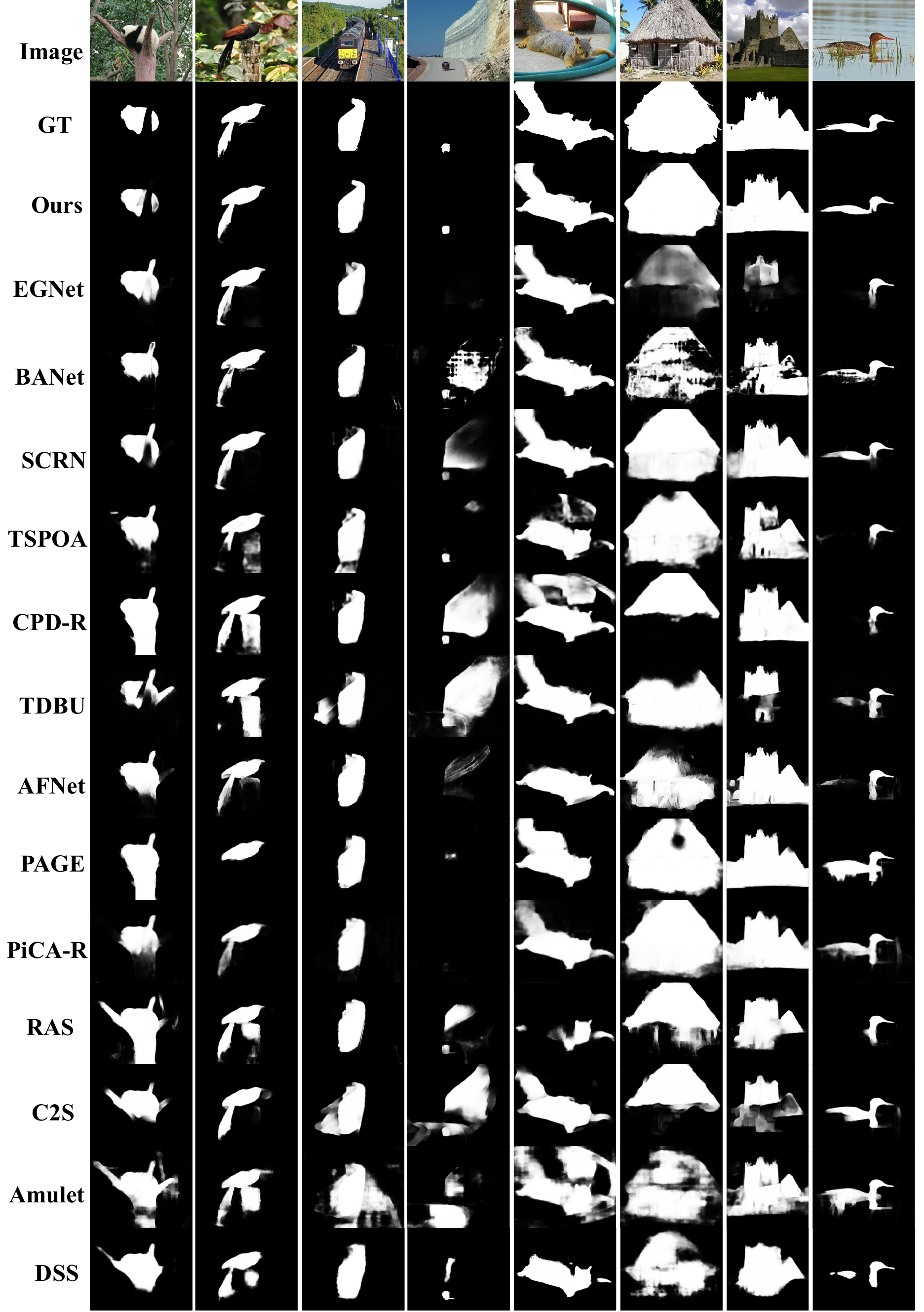}
\caption{More examples of 13 state-of-the-art methods and our approach.}
\label{fig:example}
\end{figure}

\newpage

\subsection{Quantitative Comparison}

\begin{figure}[ht!]
\centering
\includegraphics[width=1\columnwidth]{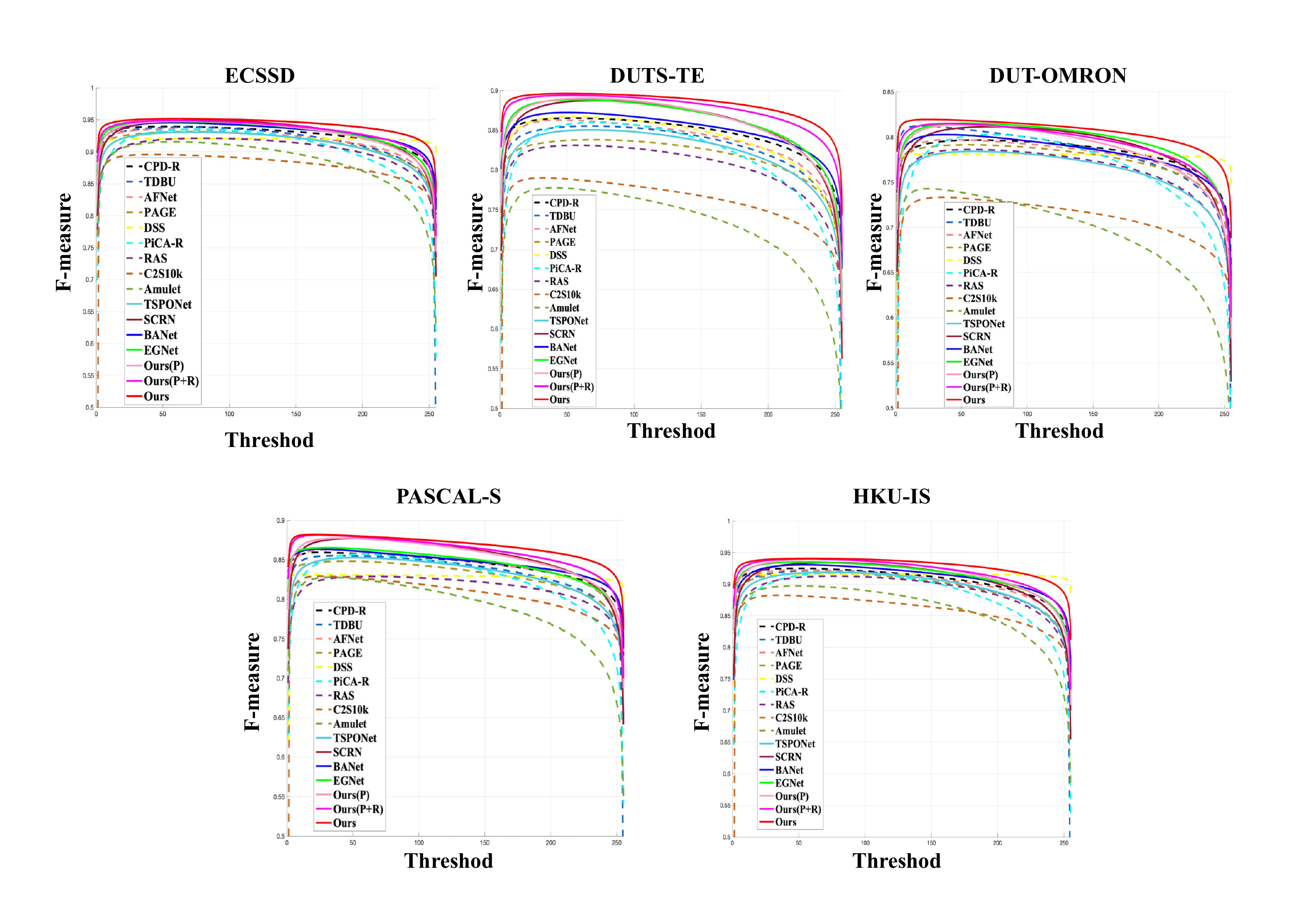}
\caption{Comparison of the F-measure curves across five benchmark datasets.}
\label{fig:example}
\end{figure}

\begin{table}[]
\caption{Performace comparison between our approach (ResNet-50) and new state-of-the-art models.}
\centering
\scalebox{0.75}{
\begin{tabular}{c|ccc|ccc|ccc|ccc|ccc}
\hline
                         & \multicolumn{3}{c|}{ECSSD}                                                                 & \multicolumn{3}{c|}{DUTS-TE}                                                                & \multicolumn{3}{c|}{DUT-OMRON}                                                             & \multicolumn{3}{c|}{PASCAL-S}                                                              & \multicolumn{3}{c}{HKU-IS}                                                                \\ \cline{2-16} 
\multirow{-2}{*}{Models} & $F_\beta$                    & $S_m$                        & MAE                          & $F_\beta$                    & $S_m$                        & MAE                          & $F_\beta$                    & $S_m$                        & MAE                          & $F_\beta$                    & $S_m$                        & MAE                          & $F_\beta$                    & $S_m$                        & MAE                          \\ \hline
SCRN(ICCV2019)           & 0.918                        & 0.927                        & 0.037                        & 0.808                        & 0.885                        & 0.04                         & 0.746                        & 0.837                        & 0.056                        & 0.827                        & 0.842                        & 0.062                        & 0.896                        & 0.916                        & 0.034                        \\
BANet(ICCV2019)          & 0.923                        & 0.924                        & 0.035                        & 0.808                        & 0.885                        & 0.04                         & 0.746                        & 0.832                        & 0.059                        & 0.823                        & 0.845                        & 0.069                        & 0.9                          & 0.913                        & 0.032                        \\
EGNet(ICCV2019)          & 0.92                         & 0.925                        & 0.037                        & 0.815                        & 0.879                        & 0.04                         & 0.755                        & 0.841                        & 0.053                        & 0.817                        & 0.846                        & 0.073                        & 0.901                        & 0.918                        & 0.031                        \\
F3N(AAAI2020)            & 0.925                        & 0.924                        & 0.033                        & 0.84                         & 0.888                        & 0.035                        & 0.766                        & 0.838                        & 0.053                        & 0.84                         & 0.855                        & 0.062                        & 0.91                         & 0.917                        & 0.028                        \\
MINet(CVPR2020)          & 0.924                        & 0.925                        & 0.033                        & 0.828                        & 0.884                        & 0.037                        & 0.755                        & 0.833                        & 0.055                        & 0.829                        & 0.85                         & 0.063                        & 0.909                        & 0.919                        & 0.029                        \\
Ours                     & {\color[HTML]{FE0000} 0.933} & {\color[HTML]{FE0000} 0.928} & {\color[HTML]{FE0000} 0.033} & {\color[HTML]{FE0000} 0.856} & {\color[HTML]{FE0000} 0.894} & {\color[HTML]{FE0000} 0.034} & {\color[HTML]{FE0000} 0.774} & {\color[HTML]{FE0000} 0.838} & {\color[HTML]{FE0000} 0.052} & {\color[HTML]{FE0000} 0.849} & {\color[HTML]{FE0000} 0.863} & {\color[HTML]{FE0000} 0.059} & {\color[HTML]{FE0000} 0.921} & {\color[HTML]{FE0000} 0.923} & {\color[HTML]{FE0000} 0.028} \\ \hline
\end{tabular}}
\label{newcomparsion}
\end{table}

\begin{table}[]
\centering
\caption{Performance on SOC of different attributes. The last row shows the whole performance on the SOC dataset.The best two results are in {\color[HTML]{FF0000} red} and {\color[HTML]{32CB00} green} fonts.}
\begin{tabular}{c|ccccc}
\hline
Attr & SCRN                         & EGNet & F3N                          & MINet                        & Ours                         \\ \hline
AC   & 0.759                        & 0.756 & {\color[HTML]{32CB00} 0.784} & {\color[HTML]{FE0000} 0.79}  & {\color[HTML]{32CB00} 0.784} \\
BO   & 0.747                        & 0.702 & 0.791                        & {\color[HTML]{32CB00} 0.813} & {\color[HTML]{FE0000} 0.814} \\
CL   & 0.766                        & 0.726 & 0.757                        & {\color[HTML]{32CB00} 0.77}  & {\color[HTML]{FE0000} 0.773} \\
HO   & 0.78                         & 0.756 & {\color[HTML]{32CB00} 0.79}  & {\color[HTML]{FE0000} 0.792} & {\color[HTML]{32CB00} 0.79}  \\
MB   & 0.719                        & 0.687 & {\color[HTML]{FE0000} 0.761} & 0.708                        & {\color[HTML]{32CB00} 0.75}  \\
OC   & {\color[HTML]{FE0000} 0.732} & 0.702 & 0.724                        & {\color[HTML]{32CB00} 0.729} & 0.725                        \\
OV   & 0.781                        & 0.764 & {\color[HTML]{FE0000} 0.793} & {\color[HTML]{32CB00} 0.785} & {\color[HTML]{32CB00} 0.785} \\
SC   & 0.709                        & 0.683 & {\color[HTML]{FE0000} 0.747} & 0.726                        & {\color[HTML]{32CB00} 0.745} \\
SO   & 0.645                        & 0.614 & {\color[HTML]{32CB00} 0.668} & 0.652                        & {\color[HTML]{FE0000} 0.68}  \\ \hline
Avg  & 0.738                        & 0.71  & {\color[HTML]{32CB00} 0.757} & 0.752                        & {\color[HTML]{FE0000} 0.761} \\ \hline
\end{tabular}
\label{soc}
\end{table}

\begin{table}[t]
\caption{Average speed (FPS) comparisons between our approach (ResNet-50) and the previous state-of-the-art methods.}
\centering
\scalebox{0.9}{
\begin{tabular}{c|p{2cm}<{\centering}p{2cm}<{\centering}p{2cm}<{\centering}p{2cm}<{\centering}p{2cm}<{\centering}p{2cm}<{\centering}}
\hline
     & Ours             & BANet            & SCRN             & AFNet            & PAGE             & CPD              \\ \hline
Size & $352 \times 352$ & $400 \times 300$ & $352 \times 352$ & $224 \times 224$ & $224 \times 224$ & $352 \times 352$ \\ \hline
FPS  & 40               & 13               & 38               & 26               & 25               & 62               \\ \hline
     & EGNet            & PiCA             & RAS              & C2SNet           & Amulet           & DSS              \\ \hline
Size & $400 \times 300$ & $224 \times 224$ & $400 \times 300$ & $400 \times 300$ & $256 \times 256$ & $224 \times 224$ \\ \hline
FPS  & 12               & 7                & 45               & 30               & 16               & 12               \\ \hline
\end{tabular}}
\label{fps}
\end{table}

F-measure curves of different methods are displayed in Fig. 2, for overall comparisons. One can observe that our approach noticeably outperforms all the other state-of-the-art methods. These observations demonstrate the efficiency and robustness of our CLASS net across various challenging datasets.

To further demonstrate the efficiency and robustness of our CLASS net, we compare our method with two new state-of-the-art methods, including F3N~\cite{DBLP:journals/corr/abs-1911-11445} and MINet~\cite{DBLP:conf/cvpr/PangZZL20}. The results are reported in Table.\ref{newcomparsion}. It can be seen that our method consistently outperforms other methods across five benchmark datasets. SOC~\cite{DBLP:conf/eccv/FanCLGHB18} is a new challenging dataset with nine attributes. In Table.\ref{soc}, we evaluate the mean F-measure score of our method in this dataset. We can see the proposed model achieves the competitive results among most of attributes and the overall score is best.

Average speed (FPS) comparisons among different methods (tested in the same environment) are also reported in Table.\ref{fps}. As can be seen, our approach is one of the fastest methods which can run in real time. 
Although there is a small gap between our method and two fastest methods CPD~\cite{DBLP:conf/cvpr/WuSH19} and RAS \cite{DBLP:conf/eccv/ChenTWH18} in fps, our method performs much better on other evaluation metrics.
This observation can further demonstrate the efficiency of our CLASS net.

\newpage

\section{Failure Cases}

\begin{figure}[ht!]
\centering
\includegraphics[scale=0.8, width=0.8\columnwidth]{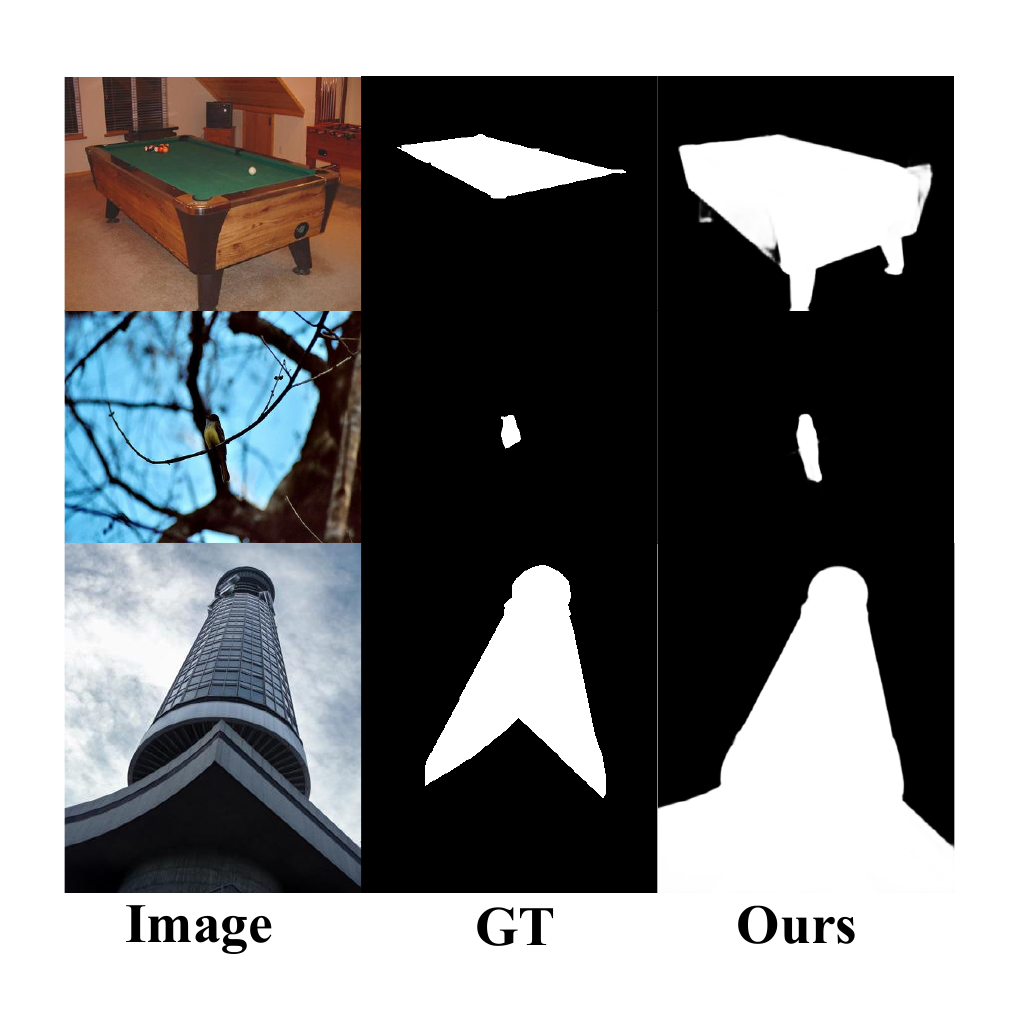}
\caption{Examples which correct the ground truth.}
\label{fig:example}
\end{figure}

\begin{figure}[ht!]
\centering
\includegraphics[scale=0.8, width=1\columnwidth]{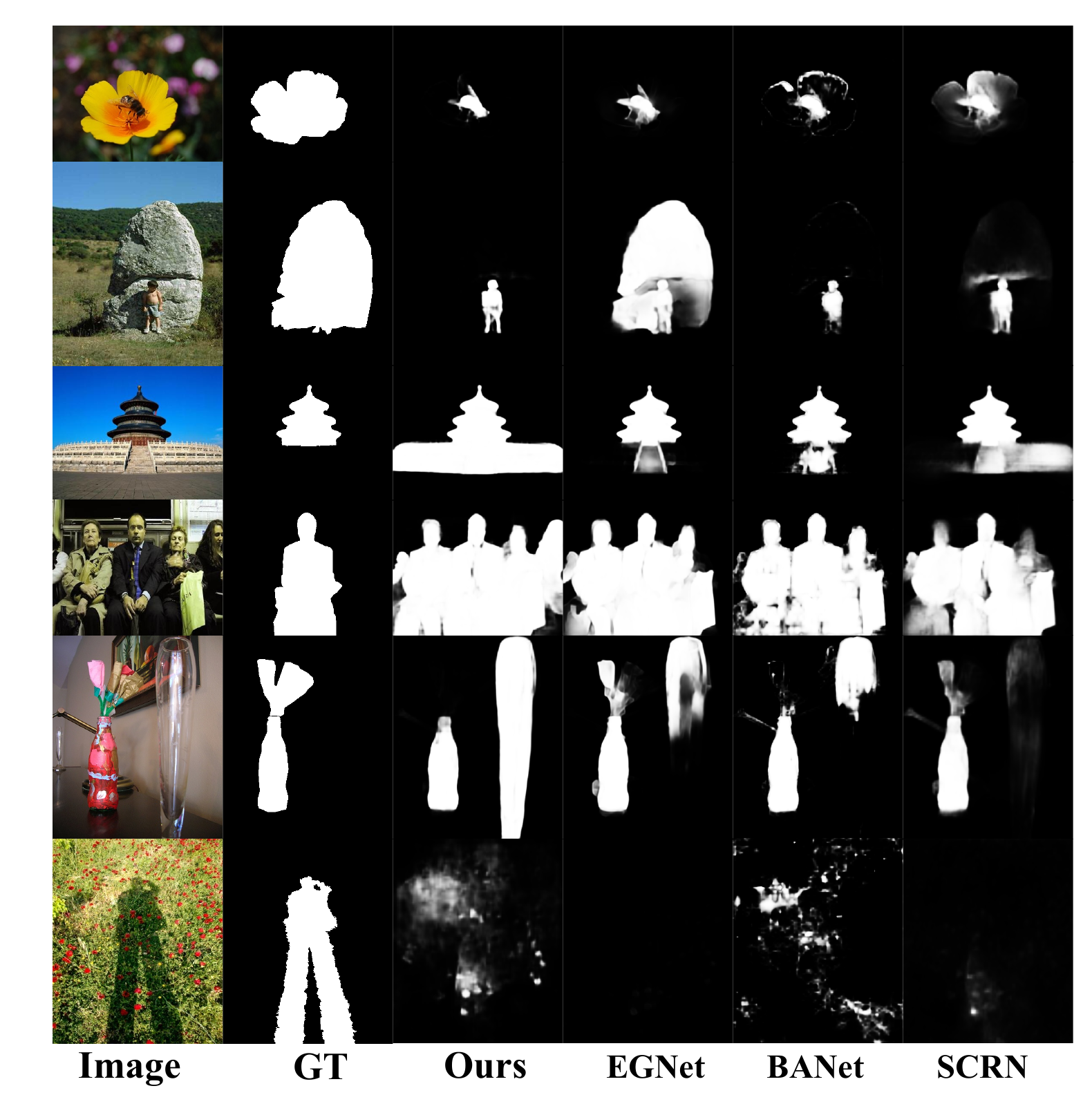}
\caption{Failure cases. }
\label{fig:example}
\end{figure}

As demonstrated, our method has achieved impressive performance in accurate salient object detection. However, there are still some cases where our detection results are inconsistent with the ground truth. 

It is noteworthy that being inconsistent with the ground truth does not mean all these cases are necessarily inferior results. As shown in Fig. 3, some of our results can even correct the errors in the ground truth by maintaining the wholeness of salient objects. 

Besides, we show several typical failure cases of our
method in Fig. 4. From the row of 1 and 2 in Fig. 4, we can observe that in some controversial scenes, our method tend to only segment the top salient object in the image. In the row of 3 and 4, our method labels all relevant regions of the salient objects while the ground truth only labels parts of the salient objects. This situation can be caused by the proposed cross-level attention mechanism, which is designed to keep the wholeness of the salient objects. In the fifth row, our method fails to detect the subjective salient object. In the last row of Fig. 4, our method cannot detect the true salient object, that can be caused by the bias of the training data. In most training images, shadows are not labeled as salient object. It is worth noting that these failure cases are also hard to most of the other state-of-the-art methods. Therefore, there is still a room for the improvement of our CLASS net.

\newpage

\section{More Analyses of the proposed CLA and CLS}
\subsection{Analysis of Cross-level Attention }
To further demonstrate the novelty of the proposed cross-level attention, we compare our attention module with the common non-local attention \cite{DBLP:conf/cvpr/0004GGH18}, which relies on a single layer feature. The quantitative results are shown in Table 4. We first remove all attention module in the proposed model as a baseline. Then we replace the cross-level attention module with the common non-local attention. As can be seen, using the common non-local attention can improve the performance of baseline. However, the common non-local models \cite{DBLP:conf/cvpr/0004GGH18} rely on a single layer feature,  they cannot leverage the advantages of features in different levels to  capture sufficient long range dependencies. The proposed cross-level attention outperforms the common non-local attention and achieves the best results on all datasets.

\begin{table}[]
\caption{Performance comparison of different attention settings. The \textbf{Baseline} here refers to without any attention module. The \textbf{Common Non-Local} means we use common non-local module to replace the proposed cross-level attention module.  }
\centering
\scalebox{0.72}{
\begin{tabular}{c|ccc|ccc|ccc|ccc|ccc}
\hline
                                                   & \multicolumn{3}{c|}{ECSSD}                                                                     & \multicolumn{3}{c|}{DUTS-TE}                                                                   & \multicolumn{3}{c|}{DUT-OMRON}                                                                 & \multicolumn{3}{c|}{PASCAL-S}                                                                  & \multicolumn{3}{c}{HKU-IS}                                                                    \\ \cline{2-16} 
\multirow{-2}{*}{Conﬁgurations}                    & \multicolumn{1}{c|}{$F_{\beta}$} & \multicolumn{1}{c|}{$S_m$}   & MAE                          & \multicolumn{1}{c|}{$F_{\beta}$} & \multicolumn{1}{c|}{$S_m$}   & MAE                          & \multicolumn{1}{c|}{$F_{\beta}$} & \multicolumn{1}{c|}{$S_m$}   & MAE                          & \multicolumn{1}{c|}{$F_{\beta}$} & \multicolumn{1}{c|}{$S_m$}   & MAE                          & \multicolumn{1}{c|}{$F_{\beta}$} & \multicolumn{1}{c|}{$S_m$}   & MAE                          \\ \hline
Baseline                                           & 0.923                            & 0.922                        & 0.037                        & 0.845                            & 0.888                        & 0.036                        & 0.767                            & 0.839                        & 0.055                        & 0.842                            & 0.856                        & 0.062                        & 0.910                            & 0.915                        & 0.031                        \\
Common Non-Local                                   & 0.928                            & 0.926                        & 0.035                        & 0.849                            & 0.889                        & 0.036                        & 0.769                            & 0.839                        & 0.055                        & 0.844                            & 0.856                        & 0.062                        & 0.915                            & 0.918                        & 0.029                        \\
Cross-Level Attention(Ours) &0.933     & 0.928 & 0.033 & 0.856    &0.894 &0.034 &0.774     &0.842 & 0.052 &0.849     &0.863 & 0.059 &0.921     & 0.923 &0.028 \\ \hline
\end{tabular}}
\end{table}

\begin{figure}[ht!]
\centering
\includegraphics[width=1\columnwidth]{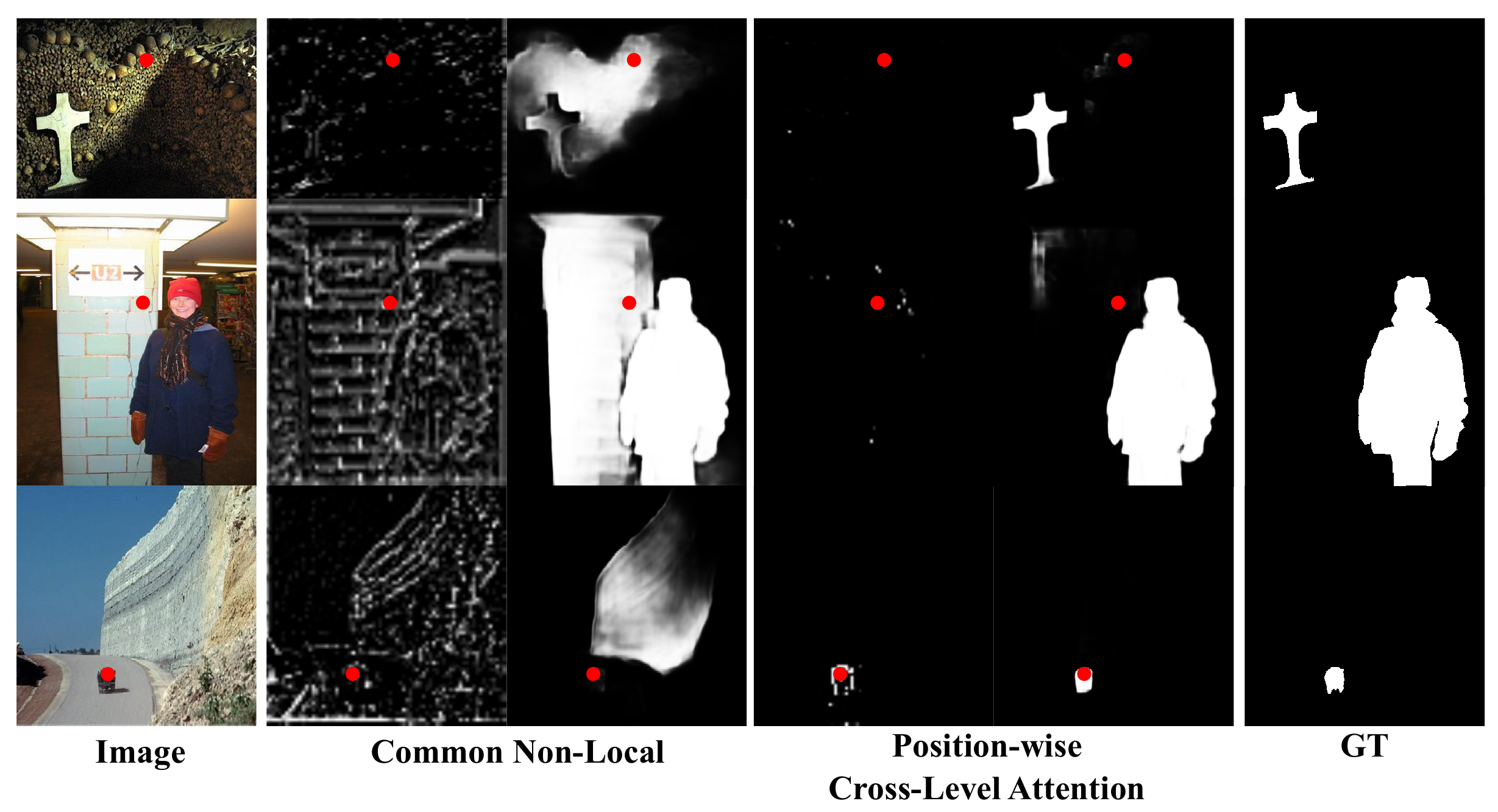}
\caption{Visual comparison of position-wise non-local attention.}
\label{fig:example}
\end{figure}

We also provide some visualization results of attention module for qualitative comparison. For position attention, since the overall attention map is calculated on all positions, there is an corresponding sub-attention map for each specific point in the image. In Fig. 5, for each input image, we select a point (marked by red dot) and show its corresponding sub-attention map as well as the saliency result of the image respectively. We observe that for some ``salient-like'' positions, common non-local sub-attention map provides a strong connection with real salient regions, which can lead a wrong prediction in these positions. While in the proposed cross-level sub-attention maps, these  positions almost have no dependencies on the real salient regions. For the position in the real salient region(third row), the cross-level sub-attention map only highlights the real salient object while the common non-local sub-attention map highlights the interfering region. These visual comparisons show our position-wise cross-level attention can better locate the salient objects and suppress the non-salient regions.

\begin{figure}[ht!]
\centering
\includegraphics[width=1\columnwidth]{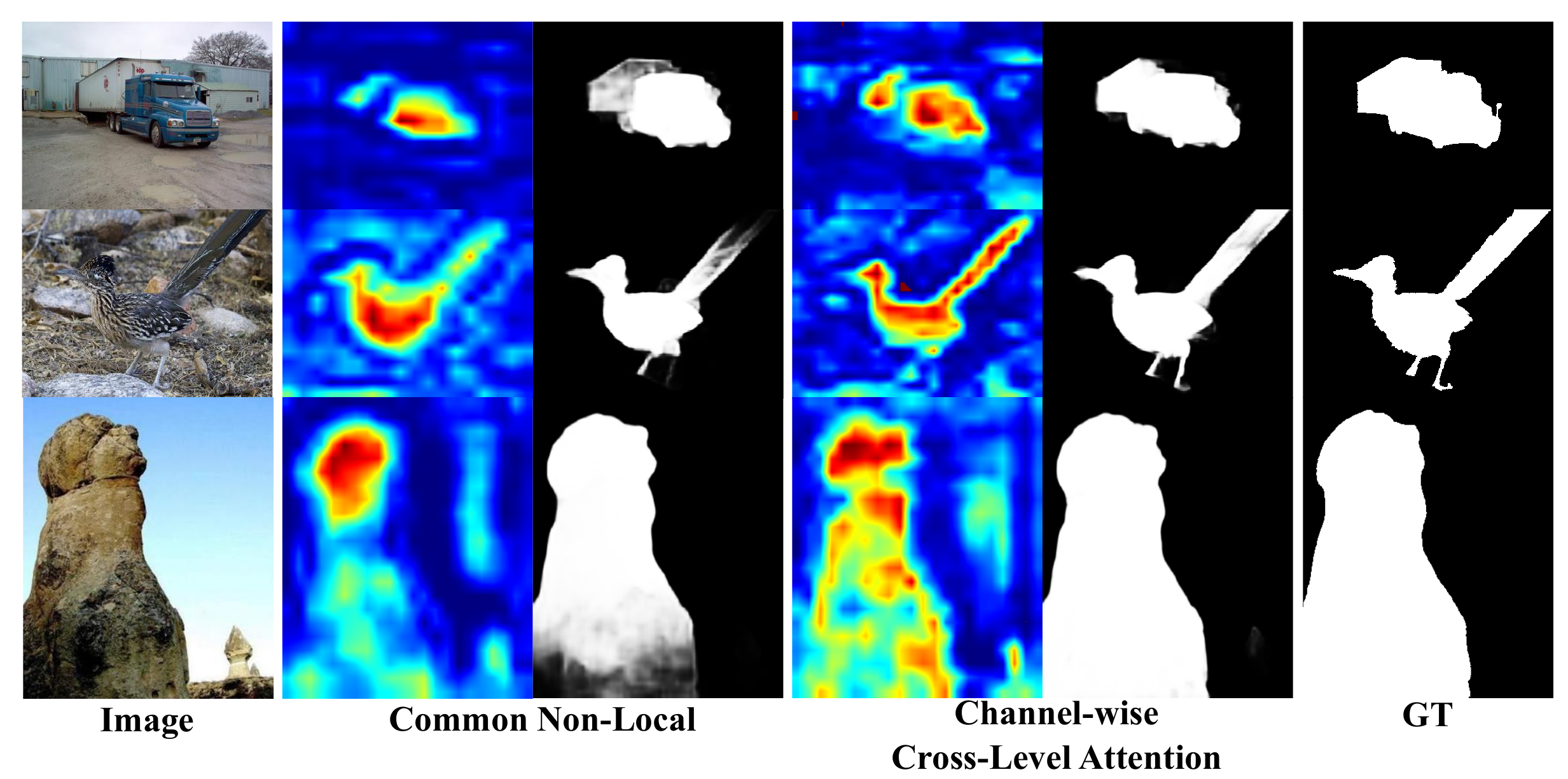}
\caption{Visual comparison of channel-wise non-local attention.}
\label{fig:example}
\end{figure}

For channel attention, it is hard to give comprehensible visualization about the attention map directly. Instead, we fuse the most attended channels provided by common non-local channel attention module and the proposed cross-level channel attention module to see whether they highlight clear semantic areas. In Fig. 6, we can find that the response of salient semantic becomes more noticeable after two kinds channel attention module enhances. However, our cross-level channel-wise attention can better keep the wholeness of salient by highlighting the regions which have different visual appearances (different color, texture and luminance) with the main salient object. 

In short, these visualizations further demonstrate the necessity of capturing cross-level long-range dependencies for improving feature representation in SOD.

\subsection{Analysis of Cross-level Supervision}
In Fig. 7, we provide a visual comparison with different supervision settings. As can be seen, by adding the region-level supervision, our model can better maintain the structural details and boundaries of the salient objects. When add the object-level supervision, our model can highlight the salient object more uniformly. The F-measure curves of different supervision settings are also provided in Fig. 2.  these visualizations further shows the effectiveness of our proposed cross-level supervision.

\begin{figure}[ht!]
\centering
\includegraphics[width=1\columnwidth]{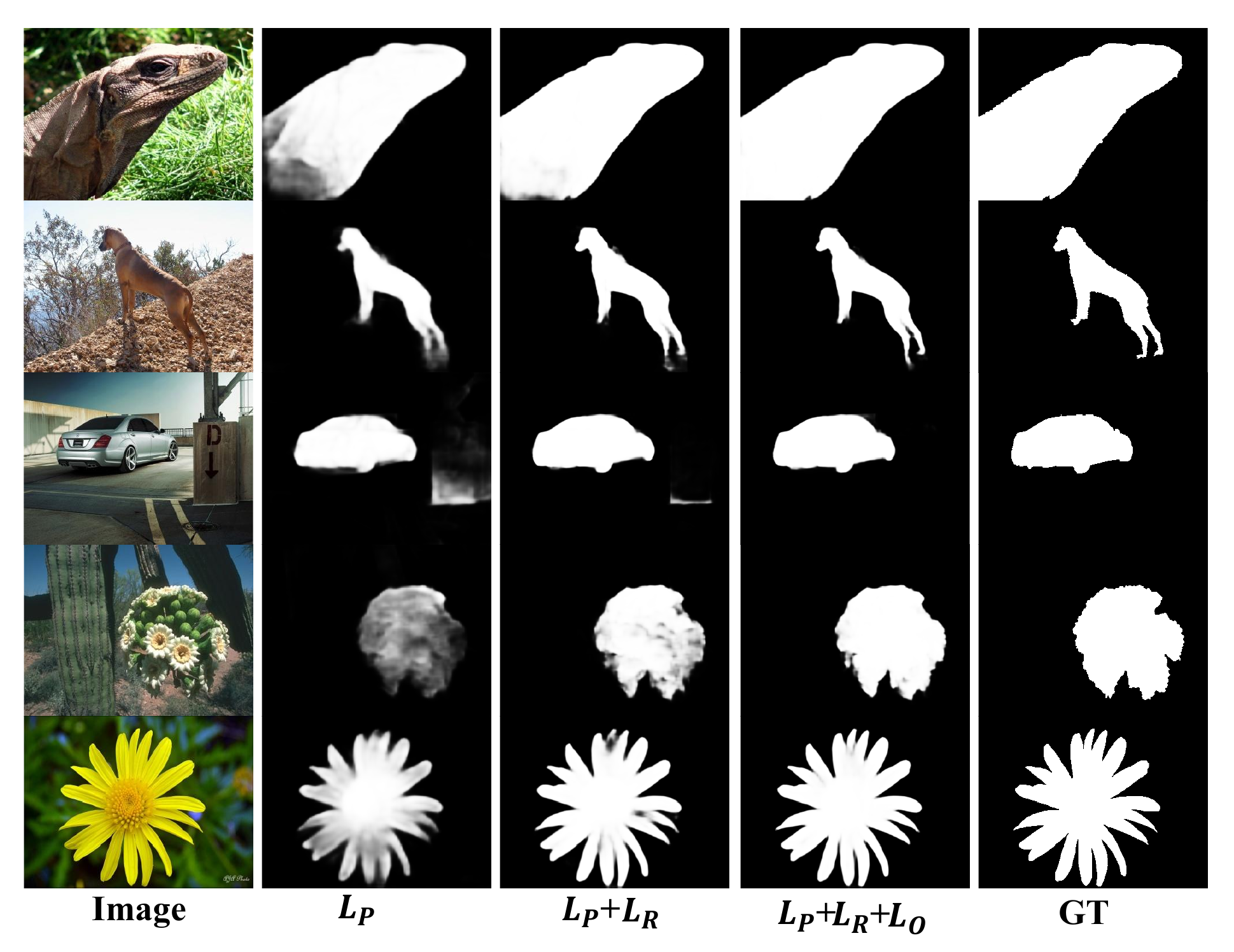}
\caption{Visual comparison of different supervision settings.}
\label{fig:example}
\end{figure}

\clearpage

\bibliographystyle{splncs}
\bibliography{egbib}

\end{document}